\documentclass[letterpaper,11pt]{article}
\PassOptionsToPackage{table}{xcolor}

\usepackage[utf8]{inputenc}
\usepackage[T1]{fontenc}
\usepackage{times}
\usepackage[margin=1in,top=0.45in]{geometry}
\usepackage{microtype}
\PassOptionsToPackage{numbers, sort&compress}{natbib} 

\usepackage{amsmath,amssymb,amsfonts,amsthm}
\usepackage{mathrsfs}
\usepackage{dsfont}

\usepackage{graphicx}
\usepackage[export]{adjustbox}
\usepackage{tabularx}
\usepackage{booktabs}
\usepackage{multirow}
\usepackage{makecell}
\usepackage{subfigure}
\usepackage{longtable}
\usepackage{wrapfig}
\usepackage{algorithm}
\usepackage{algpseudocode}

\usepackage{enumitem}
\usepackage{xspace}
\usepackage{blindtext}
\usepackage[most]{tcolorbox}
\usepackage{xcolor}
\usepackage{todonotes}
\usepackage{amsmath}
\usepackage{enumitem}

\usepackage{booktabs} 
\usepackage{siunitx}
\usepackage{etoolbox}
\usepackage{algorithm}
\usepackage{algpseudocode}
\usepackage{multirow}
\usepackage{multicol}
\usepackage{graphicx}
\usepackage{xspace}
\usepackage{wrapfig}

\usepackage[table]{xcolor}
\definecolor{lightgray}{gray}{0.92}
\definecolor{resultgreen}{HTML}{15803D}
\definecolor{resultred}{HTML}{B91C1C}
\newcommand{\resultgain}[1]{\textcolor{resultgreen}{#1}}
\newcommand{\resultloss}[1]{\textcolor{resultred}{#1}}

\DeclareRobustCommand{\shadetext}[1]{%
\begingroup
\setlength{\fboxsep}{1.2pt}%
\colorbox{lightgray}{#1}%
\endgroup
}

\newcommand{\method}{UniSD\xspace}
\newcommand{\methodfull}{UniSD$^\ast$\xspace}
\newcommand{\App}{Appendix}

\makeatletter
\newcommand{\widefigurewidth}{%
  \dimexpr\if@twocolumn\linewidth\else0.6\linewidth\fi\relax
}
\makeatother

\newcommand{\singlefigurewidth}{0.96\linewidth}

\newif\ifmanualvspace

\manualvspacetrue

\newcommand{\myvspace}[1]{%
  \ifmanualvspace
    \vspace{#1}%
  \fi
}

\makeatother

\manualvspacefalse
\renewcommand{\singlefigurewidth}{0.6\linewidth}

\usepackage{natbib}
\usepackage{url}
\usepackage[colorlinks=true,linkcolor=blue!50!black,citecolor=blue!50!black,urlcolor=blue!60!black]{hyperref}
\usepackage{fontawesome5}

\hypersetup{
  pdftitle={UniSD: Towards A Unified Self-Distillation Framework for Large Language Models},
  pdfauthor={Yiqiao Jin, Yiyang Wang, Lucheng Fu, Yijia Xiao, Yinyi Luo, Haoxin Liu, B. Aditya Prakash, Chao Zhang, Josiah Hester, Jindong Wang, Srijan Kumar}
}

\theoremstyle{definition}

\renewcommand{\thefootnote}{\fnsymbol{footnote}}

\setlist{leftmargin=5mm}

\definecolor{abstractbg}{RGB}{230,242,250}
\definecolor{abstractborder}{RGB}{200,210,220}
\usepackage{titlesec}

\titleformat{\section}
  {\sffamily\bfseries\Large} % font style
  {\thesection}              % label
  {1em}                      % spacing
  {}                         % before-code

\renewenvironment{abstract}
{
\begin{center}
\begin{tcolorbox}[
    colback=abstractbg,
    colframe=abstractborder,
    boxrule=0.6pt,
    arc=6pt,
    width=\textwidth,
    left=8pt,
    right=8pt,
    top=8pt,
    bottom=8pt
]
}
{
\end{tcolorbox}
\end{center}
}

\newcommand{\papertitle}{%
\sffamily\bfseries\fontsize{16}{20}\selectfont
\method: Towards A Unified Self-Distillation Framework for Large Language Models
}

\newcommand{\paperauthors}{% 
\sffamily
Yiqiao Jin$^{1\ast}$, Yiyang Wang$^{1\ast}$, Lucheng Fu$^1$, Yijia Xiao$^2$, \\
Yinyi Luo$^3$, Haoxin Liu$^1$, B. Aditya Prakash$^1$, Chao Zhang$^1$, \\
Josiah Hester$^1$, Jindong Wang$^{4\dag}$, Srijan Kumar$^{1\dag}$
}

\newcommand{\paperdate}{$^1$Georgia Institute of Technology \quad $^2$University of California, Los Angeles \\ \quad $^3$Carnegie Mellon University \quad$^4$William \& Mary \\
\faIcon{globe} Website: \url{https://unifiedsd.github.io/} \\
\faIcon{github} GitHub: \url{https://github.com/Ahren09/UniSD}
}

\begin{document}

\thispagestyle{empty}

% Logo at top-left
\noindent
\includegraphics[height=.6cm]{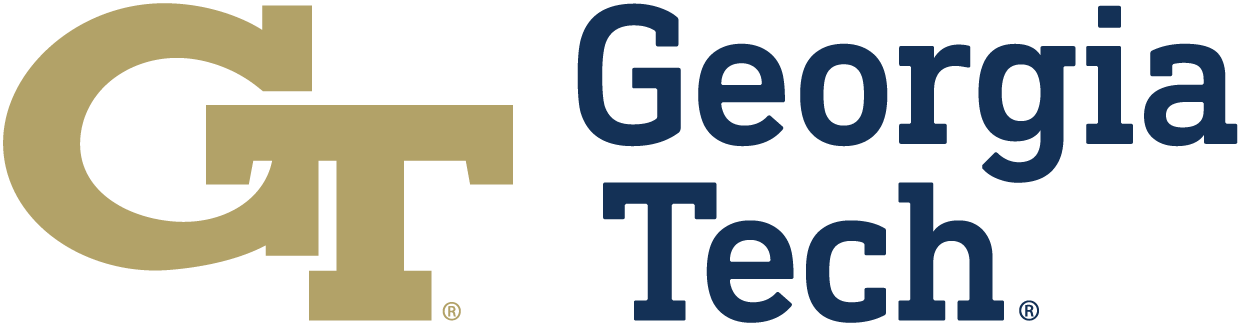} ~
\includegraphics[height=.6cm]{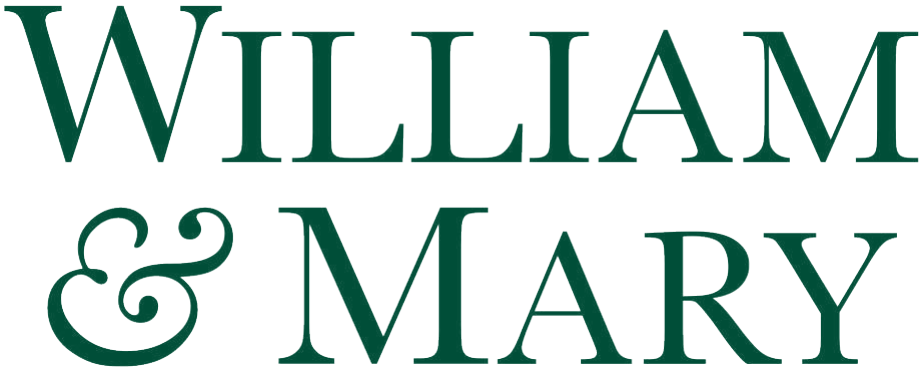} ~
\includegraphics[height=.6cm]{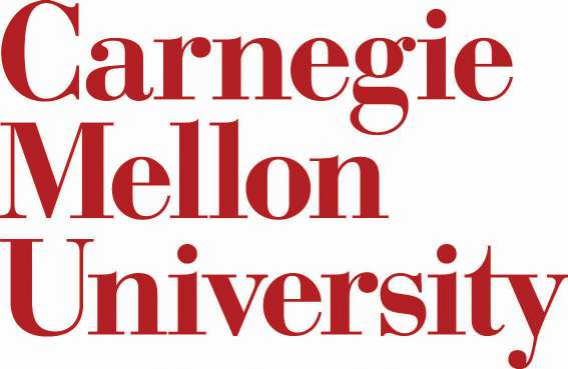} ~
\includegraphics[height=.5cm]{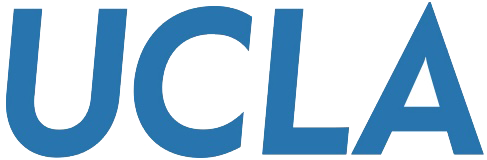} ~

% Horizontal line immediately below
\noindent\rule{\textwidth}{0.8pt}

\vspace{0.35cm}

% Title block
\begin{center}
    {\papertitle\par}
    \vspace{0.35cm}
    {\large \paperauthors\par}
    \vspace{0.2cm}
    {\normalsize \paperdate\par}
\end{center}

\footnotetext[1]{Equal contribution. Contact: yjin328@gatech.edu, ywang3420@gatech.edu.}
\footnotetext[2]{Corresponding authors: jdw@wm.edu, srijan@gatech.edu.}

\setcounter{footnote}{0}
\renewcommand{\thefootnote}{\arabic{footnote}}
\suppressfloats[t]

\begin{abstract}
Self-distillation (SD) offers a promising path for adapting large language models (LLMs) without relying on stronger external teachers.
In autoregressive LLMs, self-generated trajectories are free-form and correctness is task-dependent, motivating explicit controls over the reliability and stability of self-derived supervision.
% Self-derived supervision can be noisy and unstable.
Existing methods mainly examine isolated design choices, leaving open when self-distillation helps, which controls make it reliable, and how different mechanisms can be combined.
We introduce \textbf{\method}, a \underline{\textbf{Uni}}fied framework for systematically studying \underline{\textbf{S}}elf-\underline{\textbf{D}}istillation.
\method formulates self-distillation as reliability-aware on-policy distillation: self-derived teacher views score student-sampled trajectories, and the resulting distillation objectives update the model parameters.
For controlled comparison, \method groups the mechanisms by three functional roles: supervision reliability, representation alignment, and training stability. 
Across six benchmarks and six models from three model families, \method yields several empirical findings. Static imitation yields limited, task-dependent gains, while on-policy distillation provides a stronger starting point. Agreement trades peak performance against robustness and varies with context construction. 
Guided by these insights, we construct \methodfull, an integrated pipeline with the highest average performance of 73.3\%, exceeding Raw by 5.4 points and the strongest evaluated distillation baseline by 2.8.
% Our results show that LLMs can improve both in-domain and out-of-domain without stronger external teachers, highlighting self-distillation as a practical path toward efficient and steerable model adaptation.
% Our code is at \url{https://anonymous.4open.science/r/UniSD}.
\end{abstract}

% \setcounter{footnote}{0}
% \renewcommand{\thefootnote}{\arabic{footnote}}

% \section{Introduction}

% \citet{azaria2022chatgpt} proposed to do xxx, which is used in the paper  \citep{ahuja2023closer}.
% \blindtext\footnote{abcd}

\section{Introduction}
\label{sec:intro}
% Encountered early so this double float claims an earlier page than the
% framework figure (Figures 1 and 2 must not share a page).

\begin{figure*}[t]
\centering
\includegraphics[width=\widefigurewidth]{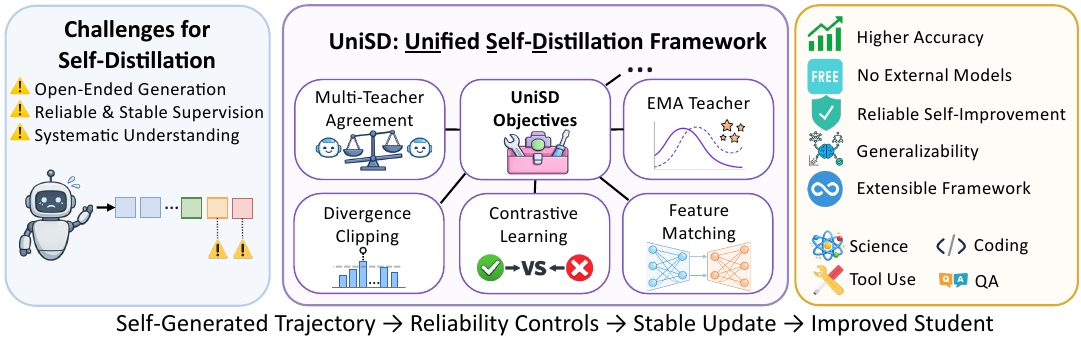}
\myvspace{-4mm}
\caption{
Overview of our controlled study of on-policy self-distillation in LLMs.
All variants---agreement, EMA stabilization, divergence clipping, contrastive learning, and feature matching---share the same on-policy sampling and optimization protocol. \methodfull combines the five controls without a stronger external teacher.
}
\label{fig:hero}
\myvspace{-3mm}
\end{figure*}

As large language models (LLMs) are adapted to new domains and tasks, supervision often comes from stronger external models through synthetic data generation~\cite{liu2023visual,taori2023alpaca,hu2024visual} or knowledge distillation~\cite{yang2025qwen3,hinton2015distilling}.
Although effective, this dependence can make adaptation costly and contingent on models restricted by access, policy, or licensing~\cite{anand2023gpt4all}; external teachers can also propagate bias or privacy-sensitive content~\cite{luo2026agentark}.
Self-distillation (SD), in which a model learns from supervision derived from itself rather than from a stronger teacher, offers an appealing alternative. 
However, self-distillation does not yield uniform improvements: training on self-derived supervision can consolidate useful capabilities but also amplify model errors.
This dual effect motivates a central research question: \emph{under what conditions does on-policy self-distillation improve LLMs?}

\myvspace{-3mm}
\paragraph{Challenges.}
On-policy self-distillation trains on free-form trajectories sampled from the current model, rather than on fixed prediction targets.
A prompt may admit multiple valid answers or reasoning paths, and each generated prefix changes the subsequent conditioning state. %~\cite{shenfeld2026self,xu2024survey,wang2025companioncast}.
Consequently, a self-generated trajectory can mix reliable and misleading decisions, while the supervision used to score it changes as training proceeds.
Effective learning therefore depends on two coupled properties: \emph{reliability}, or whether a self-derived signal should be trusted, and \emph{stability}, or whether repeated updates preserve useful behavior instead of amplifying transient errors.
Existing studies mainly explore individual strategies, but do not reveal which controls are responsible and how their benefits depend on the setting. 

\paragraph{This Work.}
We propose \method, a \underline{\textbf{Uni}}fied framework for systematically studying \underline{\textbf{S}}elf-\underline{\textbf{D}}istillation in LLMs. 
Using \method, we conduct a comprehensive empirical study of various reliability and stability controls under a shared on-policy distillation protocol.
For every variant, the student samples a completion from its current policy, self-derived teacher views score that fixed trajectory, and the resulting objective updates the student using the same data, rollout procedure, and optimization settings.
The variants differ in the control under study: \emph{Multi-Teacher Agreement} tests whether cross-view consistency can identify trustworthy supervision; \emph{Token-Level Contrastive Learning} tests whether separating useful signals from plausible but incorrect alternatives improves learning; \emph{Feature Matching} tests whether representation-level transfer adds information beyond output distributions. \emph{EMA Teacher} and \emph{Divergence Clipping} test whether smoothing targets and limiting extreme token losses stabilize updates.
This shared protocol makes component effects directly comparable for controlled ablations.

% \myvspace{-1mm}
Our results show that the standalone effects of the controls vary across settings.
On-policy self-distillation improves more than static imitation overall, but its benefits remain sensitive to task structure and noisy supervision.
Agreement's benefits vary across tasks and model scales. In the separate scale study, EMA improves ScienceQA at both 0.5B and 1.5B; GPQA rankings differ.
Context construction matters more than adding more teacher views, and stronger disagreement penalties trade adaptivity for robustness. 
In the primary Qwen2.5-7B comparison, \methodfull achieves the highest six-benchmark mean: +5.4 points over Raw and +2.8 over GKD, the strongest evaluated distillation baseline. Mean gains over Raw extend across all three tested model families. 
These findings show how reliability filtering, representation transfer, and stabilization affect self-distillation.

% Anchored here (rather than at the start of Section 2) so this double float is
% encountered while page 2 is still being built and can be placed on page 3,
% on a different page from Figure 1.
\begin{figure*}[t]
\centering
\includegraphics[width=\widefigurewidth]{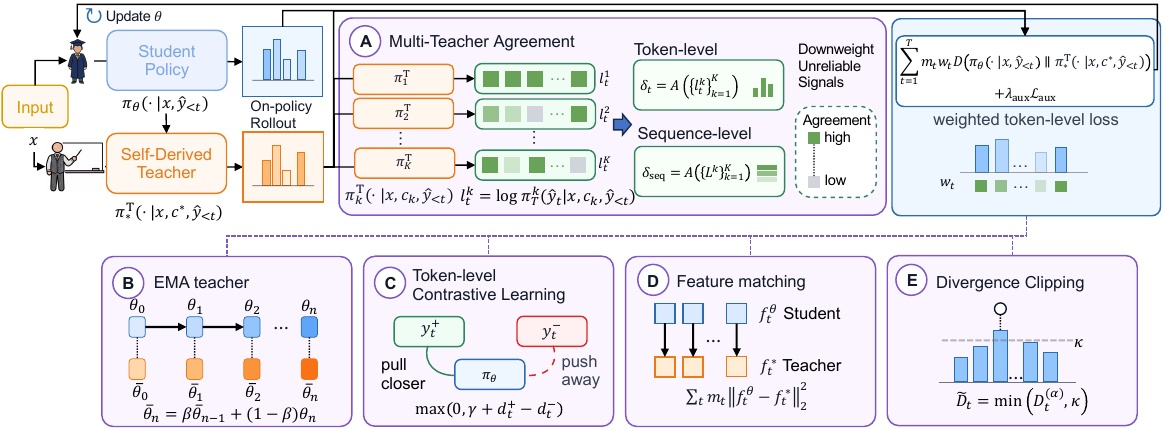}
\myvspace{-5mm}
\caption{
Variants of \method share the student-sampling and optimization procedure: Multi-Teacher Agreement, EMA Teacher, Token-Level Contrastive Learning, Feature Matching, or Divergence Clipping.
The common formulation supports controlled component ablations and combinations.
}
\label{fig:method}
\myvspace{-2mm}
\end{figure*}

\paragraph{Contributions.} Our contributions are as follows.
\myvspace{-3mm}
\begin{itemize}[leftmargin=1em]
    \item \method compares self-distillation mechanisms in a shared on-policy framework with matched data, rollouts, and optimization.
    \myvspace{-2mm}
    \item Across six benchmarks and six models, we characterize how agreement granularity and context construction, temporal averaging, contrastive supervision, feature alignment, and divergence clipping affect task performance and retention.
    \myvspace{-2mm}
    \item \methodfull combines these controls, improves 15 of 18 model--dataset pairs over Raw, and leads the primary comparison by six-task mean.
\end{itemize}

\section{Method}
\label{sec:method}

% We present \method, a unified framework that organizes different strategies into a single reliability-aware on-policy distillation pipeline.

\subsection{Self-Distillation in Autoregressive LLMs}
\label{sec:preliminary}
% \todo{self-distillation has been studied in DL, but not in LLMs, cite paper, emphasize that it is different
% 1. Correctness is underdetermined and task-dependent.
% 2. Expansive trajectory space and error propagation.
% }
We study \emph{self-distillation} in \emph{autoregressive LLMs}, where the model improves using supervision derived from its own behavior rather than from stronger external teachers~\cite{shenfeld2026self, xu2024survey}. 
% Unlike classical knowledge distillation settings, autoregressive LLMs learn over open-ended trajectories. Correctness is often task-dependent due to the open-ended nature of questions. 
As discussed in \S\ref{sec:intro}, the task is challenging because LLM generations are open-ended and the resulting self-distillation signals can be unstable. 
Effective self-distillation must therefore select useful self-distillation signals while estimating when each signal is trustworthy. 
% Effective self-distillation must therefore distinguish reliable supervision from noisy signals, such as plausible but incorrect rationales, misleading demonstrations, and locally unreliable tokens. 
%  autoregressive LLMs induce an expansive trajectory space.
% Moreover, as each generated prefix defines a new state, small early errors can have a cascading effects, propagating the errors and causing sequence-level noise.
% The model must filter reliable signals. 
% \emph{2) Expansive trajectory space and error propagation.}
% These properties make reliability a central challenge: self-distillation must decide not only what signal to imitate, but also when that signal is trustworthy. 
% Classical self-distillation usually transfers soft labels over a fixed dataset with a limited set of prediction targets. In contrast, autoregressive LLMs can visit a much larger space of states because each generated prefix defines a new conditioning context. 
% Unlike classical knowledge distillation, which transfers information from a fixed high-capacity teacher to a smaller student, we focus on \emph{self-improvement}~\cite{xu2024survey}. 
Let $\pi_\theta$ denote the student policy. 
Given an input $x$, 
the student samples an on-policy completion $\hat{y} = (\hat{y}_1,\dots,\hat{y}_T) \sim \pi_\theta(\cdot \mid x)$. 
% Different from supervised fine-tuning, 
% Self-distillation constructs a reliable teacher signal. 
Self-distillation supervises this trajectory with a primary teacher $\pi_{*}^{\mathrm T}(\cdot \mid x,c,\hat{y}_{<t})$, while auxiliary teachers estimate the reliability of the target. 
Here, $c$ is the privileged demonstration for the current training example, such as the gold answer and rationale or reference tool calls. The student receives only $x$.
% For each completion token, we define a token-level divergence
% \begin{equation}
% \mathcal D_t
% =
% D\!\left(
% \pi_\theta(\cdot \mid x,\hat{y}_{<t}),
% \pi^{\mathrm T}(\cdot \mid x,\hat{y}_{<t})
% \right),
% \end{equation}
We train on \emph{on-policy} student trajectories:
\myvspace{-1mm}
% We use a teacher policy $\pi^{\mathrm T}$ derived from the same backbone under an alternative supervision condition, such as a privileged context, a temporal state, or a positive/negative supervision role. 
% , i.e., the privileged context under which the teacher produces the supervision target. Auxiliary teachers provide additional views, such as $\pi^{\mathrm T}(\cdot \mid x, c^{k}, \hat{y}_{<t})$ for reliability estimation. 
% where each $c^{k}$ instantiates the same teacher backbone under a different supervision condition for the same input $x$, such as a different privileged context, perturbation, temporal state, or supervision role. 
\begin{equation}
\begin{split}
\mathcal{L}
=
\mathbb E_x \,
\mathbb E_{\hat y \sim \pi_\theta(\cdot \mid x)}
\Big[
\frac{1}{Z}\sum_{t=1}^{T}
m_t w_t \,\times
\\
D\!\left(
\pi_{*}^{\mathrm T}(\cdot \mid x, c, \hat{y}_{<t})
\,\|\,
\pi_\theta(\cdot \mid x, \hat{y}_{<t})
\right)
\\
+ \lambda_{\mathrm{aux}}
\mathcal L_{\mathrm{aux}}(\theta; x,\hat{y},c)
\Big].
\end{split}
\label{eq:overall}
\end{equation}
% \todo{input/signal/teacher supervision: agreement, contrast; update policy/rule: ema, clip, and contrast; feature matching}
Here, $D$ is the selected divergence (forward KL in the main \method configuration), $w_t$ a reliability weight, $m_t$ a completion-token mask, and $Z=\max(1,\sum_t m_t w_t)$. $\mathcal L_{\mathrm{aux}}$ collects auxiliary losses; representation-only matching omits $D$.
% Different self-distillation strategies correspond to different choices of teacher construction, reliability control, temporal stabilization, and auxiliary terms within the same template. 

% \input{src/alg}

\subsection{The \method Framework}
\label{sec:framework}
\method is a \underline{\textbf{Uni}}fied framework for systematically studying reliability-aware on-policy distillation in LLMs (Algorithm~\ref{alg:method}).
\method organizes the five mechanisms studied here along three operational dimensions.
These dimensions structure our controlled comparisons by the functional roles of the five components.
First, \emph{supervision reliability}: since self-derived targets can be noisy, \emph{Agreement} identifies whether the update is supported by multiple teacher views, while \emph{Token-Level Contrastive Learning} separates useful supervision from plausible but incorrect alternatives.
Second, % \emph{internal structure}:
\emph{representation alignment}:
beyond output distributions, \emph{Feature Matching} transfers internal representational structure.
% self-distillation should not be limited to output distributions; internal representations also carry transferable structure, motivating \emph{Feature Matching} as a complementary supervision space.
Third, \emph{training stability}: \emph{EMA Teacher} smooths evolving teacher signals, while \emph{Clipping} prevents rare high-divergence tokens from dominating training.
% the evolving teacher signal, and \emph{Clipping} prevents rare high-divergence tokens from dominating the update.
This grouping distinguishes whether a mechanism primarily acts on supervision signals, internal representations, or parameter updates; a mechanism may span more than one role.
\myvspace{-2mm}
\paragraph{Multi-Teacher Agreement.}
% Alternative Titles:
% Teacher Views as Reliability Probes
Self-derived supervision signals can be \emph{noisy} and \emph{context-sensitive} and can depend on how the teacher is instantiated.
% Self-derived supervision can be sensitive to how the teacher is instantiated, especially when training data are noisy or scarce. 
Inspired by the \emph{wisdom of the inner crowd}~\cite{herzog2014harnessing}, we use multiple auxiliary teachers to evaluate the same student behavior from different perspectives. 
The auxiliary teacher views serve as \emph{reliability probes} measuring the stability of teacher signals under contextual variation rather than as additional distillation targets. 
% \yq{
% % similar to how humans learn and reflect, the key to robust self-distillation is to form multiple views that can cross-check one another and to learn in proportion to their agreement. 
% Instead, they first attempt a problem under their current knowledge, inspect multiple viable solutions, understand where their reasoning diverged, weigh evidence more when multiple perspectives agree, and gradually consolidate useful updates without overwriting prior knowledge. 
% Meanwhile, Aggregating judgments from multiple internal perspectives is often more reliable than relying on a single estimate. High agreement suggests a trustworthy signal, while greater disagreement calls for more cautious updating.} 
% \yq{a common setting is that we have multiple teachers. explain that our multi-teacher is not like voting, as many people understand}
% A single teacher can provide signals that are useful but sensitive to contexts and prompts, especially when training data are noisy or scarce. 
% We therefore formulate \method as a reliability-aware on-policy distillation framework that converts this signal into a stable training target.
Given the student-sampled completion $\hat y$, we score the same trajectory under each auxiliary teacher: 
\begin{equation}
\ell_t^{k}
=
\log
\pi_{k}^{\mathrm T}
\left(
\hat{y}_t \mid x,c^{k},\hat{y}_{<t}
\right), \quad t \in [1, T].
\label{eq:aux_ll}
\end{equation}
Variation across $\{\ell_t^{k}\}$ measures sensitivity to the auxiliary contexts. % rather than variation from different generations. 
% By comparing these attempts, they can filter out unstable cues and focus on signals that remain consistent across trials. 
We estimate disagreement at two complementary granularities. 
1) Token-level agreement measures \emph{local} score variability by computing $\delta_t=A\!\left(\{\ell_t^{k}\}_{k=1}^{K}\right)$, where $A(\cdot)$ is a variability statistic, such as variance or range. %, that measures how much the teacher-view scores vary at token $t$.
2) Sequence-level agreement captures \emph{global} instability of the completion. % of the whole completion while being less sensitive to isolated token fluctuations. 
It first aggregates each teacher view as $L^{k}=\frac{\sum_{t=1}^{T} m_t \ell_t^{k}}{\sum_{t=1}^{T} m_t}$, then computes $\delta_{\mathrm{seq}} = A(\{L^{k}\}_{k=1}^{K})$.
% Intuitively, if multiple teachers assign similar scores to the same student trajectory, the resulting supervision is more likely to be reliable. Conversely, large disagreement suggests a less trustworthy signal.
In all experiments, we instantiate $A(\cdot)$ as the variance across teacher views.
To make the agreement strength $\gamma$ comparable across batches and tasks, we normalize token-level disagreement by its 95th percentile over valid completion tokens in the batch, cap it, and apply inverse-disagreement weighting:
\begin{equation}
\begin{split}
\hat{\delta}_t&=\min\!\left(\frac{\delta_t}{\max(q_{0.95}(\delta),10^{-6})},\,10\right),
\\
w_t&=\max\!\left(\frac{1}{1+\gamma\hat{\delta}_t},\,0.01\right).
\end{split}
\label{eq:relweight}
\end{equation}
For sequence-level agreement, every valid completion token shares $w_t=\max(\exp(-\gamma\,\delta_{\mathrm{seq}}),10^{-6})$.
For token-level agreement, low-variance tokens retain weights close to $1$, while higher-variance tokens receive less relative weight in Equation~\ref{eq:overall}.
The three constants are fixed in the main comparisons. Appendix~\ref{app:objective_spec} gives their rationale and a sensitivity analysis.
Auxiliary teacher views can be generated by any task-preserving perturbation that offers an alternative perspective on the same student trajectory.
We instantiate them through context variation, where each view is computed as $\pi_{k}^{\mathrm T}(\cdot \mid x, c^{k}, \hat{y}_{<t})$.
In \methodfull, each of $K=3$ retrieved views contains one solved training example, excluding the current example.
Random examples and induced instructions~\cite{honovich2023instruction} are evaluated separately (Appendix~\ref{app:supervision_construction}).
All views share one teacher model and are batched across contexts, avoiding extra model copies.
Auxiliary contexts determine reliability weights; the primary teacher supplies the distillation target. The student uses no auxiliary context at inference (Appendix~\ref{app:icl_control}).

% Provide the same backbone under different auxiliary contexts, such as using few-shot examples or induced high-level natural language instruction~\cite{honovich2023instruction} as in-context instructions. In \S\ref{sec:overall}, we explored various teacher construction strategies. 

% context $c$, we construct a set of auxiliary contexts $\mathcal{C}(x)=\{c_1,\dots,c_K\}$. 
% These auxiliary contexts are not treated as independent supervision targets.
% Instead, they are used to estimate the reliability of the teacher signal under context perturbations.
% Here, multiple strategies can be used for constructing auxiliary contexts, such as conditioning on randomly sampled demonstrations or retrieved demonstrations based on semantic relevance to the input example, or instruction-induced context where a small set of demonstrations is abstracted into a higher-level natural-language instruction and used as $c_k$ instead of raw examples. 
% Auxiliary contexts can be constructed using multiple strategies, such as from randomly sampled demonstrations, retrieved demonstrations relevant to the input, or higher-level instructions abstracted from a small set of demonstrations.

\paragraph{Temporal Stabilization with EMA Teachers.}
% \todo{shorten desc. use paragraph} 
Reliability weighting addresses whether the current teacher signal is trustworthy, % but it does not prevent the teacher itself from drifting too rapidly across training steps. 
but it does not prevent the teacher target itself from drifting across training steps. 
In self-distillation, such temporal drift can propagate transient errors or overconfident predictions into later updates. 
We therefore use an exponential moving average (EMA) teacher to provide a temporally smoothed self-derived target. 
% track the student's progress while smoothing high-variance updates.
% resulting in both stable training and superior final performance.
% A teacher constructed from the current model can be unstable, as its predictions may change rapidly across training steps due to gradient noise and minibatch variations~\cite{mandt2017stochastic}. 
Let $n$ denote the optimization step, $\theta_n$ the student parameters, and $\bar{\theta}_n$ the EMA teacher parameters. 
We update the teacher as
\begin{equation}
\bar{\theta}_n
=
\beta \bar{\theta}_{n-1} + (1-\beta) \theta_n,
\quad \beta \in [0,1].
\label{eq:ema}
\end{equation}
Between synchronization steps the teacher is fixed. It defines $\pi_{\bar{\theta}_n}(\cdot \mid x, c, \hat{y}_{<t})$, which replaces the primary teacher $\pi_{*}^{\mathrm{T}}$ in the self-distillation objective (Equation~\ref{eq:overall}). 
% \begin{equation}
% \mathcal L_{\mathrm{ema}}(\theta)
% =
% \sum_{t=1}^{T}
% w_t \,
% D_{\mathrm{KL}}\!\left(
% \pi_\theta(\cdot \mid x, \hat{y}_{<t})
% \,\|\,
% \pi_{\bar{\theta}_n}(\cdot \mid x, c, \hat{y}_{<t})
% \right).
% \end{equation}
Agreement weights current signals; EMA smooths targets over time.
% This stabilizes self-distillation by smoothing abrupt teacher changes over training. Agreement and EMA are complementary: agreement decides whether a signal is reliable at the current step, while EMA stabilizes how the teacher evolves across steps.

\paragraph{Token-level Contrastive Learning.}
% Alternative Names:
% Contrastive Teacher Regularization
% Negative-Aware Self-Distillation
% Contrastive Demonstration Distillation
% \paragraph{Contrastive Teacher Construction}
Robust self-distillation should not only reinforce reliable teacher signals but also contrast them with plausible but incorrect alternatives. 
This is especially important when positive and negative demonstrations share substantial surface structure, such as code solutions that differ only in key implementation details. 
% This distinction is important in settings such as code generation, where positive and negative demonstrations may share the same function signature, imports, and return structure while differing only in key implementation details. 
We therefore introduce a margin-based token-level contrastive objective. 
% a contrastive auxiliary objective that compares the student trajectory against teacher signals conditioned on positive and negative supervision examples. 
Let $y^{+}$ denote positive supervision and $y^{-}$ a candidate negative answer or rationale. 
We construct $y^{-}$ from cached candidates generated by the same backbone, prompted to give an incorrect option, buggy program, or incorrect tool call. One candidate is selected per training example.
Appendix~\ref{app:supervision_construction} specifies this pipeline.
% We therefore compare a positive teacher distribution $\pi^{\mathrm T}(\cdot \mid x, c^{+}, \hat{y}_{<t})$ and a negative teacher distribution $\pi^{\mathrm T}(\cdot \mid x, c^{-}, \hat{y}_{<t})$ from the same backbone under a positive condition $c^{+}$ and a corrupted condition $c^{-}$, respectively. 
% Standard contrastive objective such as InfoNCE~\cite{oord2018representation} over sequence representations can be overly coarse for autoregressive distillation: positive and hard negative responses often share substantial surface structure while differing only in a few critical local decisions.

Given an on-policy student completion $\hat{y}=(\hat{y}_1,\dots,\hat{y}_T)$, we score the same trajectory under the student distribution and two teacher distributions conditioned on $y^{+}$ and $y^{-}$, respectively, i.e., $\ell_t^{\theta}=\log\pi_\theta(\hat{y}_t\mid x,\hat{y}_{<t})$ and $\ell_t^{\pm}=\log\pi^{\mathrm T}(\hat{y}_t\mid x,y^{\pm},\hat{y}_{<t})$, and optimize:
\begin{equation}
\mathcal L_{\mathrm{aux}}
=
\frac{\sum_{t=1}^{T}m_t\max(0,\gamma_{\mathrm m}+d_t^{+}-d_t^{-})}{\sum_{t=1}^{T}m_t},
\label{eq:contrast}
\end{equation}
where $d_t^{\pm}=|\ell_t^{\theta}-\ell_t^{\pm}|$ measure token-level distances to the positive- and negative-conditioned teacher signals, respectively. $m_t \in \{0, 1\}$ masks completion tokens and $\gamma_{\mathrm m}$ is the margin. The contrastive condition is $c=(y^+,y^-)$.
On the sampled trajectory, this loss updates the student parameters to reduce the token-level distance to the positive-conditioned teacher relative to the negative-conditioned teacher.
% Compared with a standard InfoNCE-style objective, it is better aligned with autoregressive distillation, as it preserves token-level supervision and directly models how corrupted demonstrations alter next-token preferences. 

\paragraph{Feature Matching.}
Token-level distillation aligns output distributions without directly constraining internal features.
We therefore add an optional feature-matching term that regularizes selected student features toward teacher features, such as hidden states, layer-wise representations~\cite{liang2023less}, self-attention relations or attention-derived features~\cite{wang2021minilmv2}, or other task-relevant internal signals. 
For the same on-policy completion $\hat{y}=(\hat{y}_1,\dots,\hat{y}_T)$, we extract student and teacher features at matching token positions.
Let $\mathbf f_t^\theta$ and $\mathbf f_t^\ast$ denote the selected features at token $t$. 
We optimize:
\begin{equation}
\mathcal L_{\mathrm{feat}}
=
\frac{\sum_{t=1}^{T}m_t\|\mathbf f_t^\theta-\mathbf f_t^\ast\|_2^2}{d\sum_{t=1}^{T}m_t},
\label{eq:feat}
\end{equation}
where $d$ is the hidden-state width. Match (Repr.) uses this hidden-state MSE alone; Match (Joint) and \methodfull add it to logit-distribution distillation.

\paragraph{Divergence Clipping.}
% \todo{can put some descriptions into appendix}
Rare high-divergence tokens can dominate optimization. 
We therefore clip each scalar token-level divergence after reducing over the vocabulary and before applying reliability weights. 
% Let $\pi_{\theta}(\cdot \mid x,\hat{y}_{<t})$ and $\pi^{\mathrm T\ast}(\cdot \mid x,c,\hat{y}_{<t})$ denote the student and primary-teacher distributions at completion position $t$. 
For $0<\alpha<1$, the weighted Jensen--Shannon divergence is
% we use a weighted Jensen--Shannon divergence 
\begin{align}
\mathcal D_t^{(\alpha)}
&=
\alpha D\!\left(
\pi_{*}^{\mathrm T}(\cdot \mid x,c,\hat{y}_{<t})
\,\|\, M_t
\right) \notag\\
&\quad+
(1-\alpha)D\!\left(
\pi_{\theta}(\cdot \mid x,\hat{y}_{<t})
\,\|\, M_t
\right), \label{eq:jsd} \\
M_t
&=
(1-\alpha)\pi_{\theta}(\cdot \mid x,\hat{y}_{<t}) \notag\\
&\quad+
\alpha\pi_{\ast}^{\mathrm T}(\cdot \mid x,c,\hat{y}_{<t}),
\end{align}
where $D(\cdot\,\|\,\cdot)$ denotes KL divergence. 
We additionally support forward- and reverse-KL objectives as separate endpoint-style alternatives to the weighted JSD objective. 
% We additionally support KL-style endpoint objectives as separate special cases, corresponding to teacher-to-student ($\alpha=0$) and student-to-teacher ($\alpha=1$) KL losses.
We then cap the scalar divergence as
$\widetilde{\mathcal D}_t=\min(\mathcal D_t,\kappa)$, where $\mathcal D_t$ is the selected divergence and $\kappa$ the threshold. 
% At the endpoints, the objective reduces to KL-style distillation losses. 
% $\alpha=0$ and $\alpha=1$ correspond to teacher-to-student \emph{forward KL} and student-to-teacher \emph{reverse KL}, respectively. 
With agreement weights $w_t$, the clipped distillation objective is 
% and $m_t$ is the completion-token mask. The case $\alpha=0.5$ gives the standard symmetric JSD. 
\begin{equation}
\mathcal L_{\mathrm{clip}}
=
\frac{
\sum_{t=1}^{T} m_t\, w_t\, \widetilde{\mathcal D}_t
}{
\max(1,\sum_{t=1}^{T} m_t\, w_t)
},
\label{eq:clip_obj}
\end{equation}
where $m_t$ denotes the completion-token loss mask. 

\subsection{\texorpdfstring{\methodfull}{UniSD*}: An Integrated Pipeline}
We instantiate \methodfull as a unified pipeline that integrates all objectives (\S\ref{sec:framework}). 
From the \emph{supervision} perspective, multi-teacher agreement and token-level contrastive learning select reliable self-derived signals and suppress plausible but incorrect alternatives. 
From the \emph{representation} perspective, feature matching transfers internal structure beyond output distributions. 
From the \emph{optimization} perspective, the EMA teacher and divergence clipping stabilize learning under noisy on-policy trajectories. 
These components combine signal selection, representation alignment, temporal smoothing, and loss stabilization within the same reliability-aware on-policy distillation loop.
The exact final objective of \methodfull, together with the per-variant instantiation of Equation~\ref{eq:overall} used in our ablations, is given in Appendix~\ref{app:objective_spec}.

\section{Evaluation}
\label{sec:eval}
\myvspace{-1mm}
% table* == table in one-column layouts; spans both columns in the ACL build.
\begin{table*}[t]
\centering

\myvspace{-.1in}
% 调整列间距，使表格更紧凑
% \setlength{\tabcolsep}{4pt}
\footnotesize
\setlength{\tabcolsep}{2pt}
\renewcommand{\arraystretch}{1.08}
\definecolor{shading}{gray}{0.95}
\begin{tabular}{
>{\raggedright\arraybackslash}p{.225\textwidth}
*{4}{>{\centering\arraybackslash}p{.073\textwidth}}
>{\columncolor{shading}\centering\arraybackslash}p{.073\textwidth}
>{\columncolor{shading}\centering\arraybackslash}p{.073\textwidth}
>{\centering\arraybackslash}p{.073\textwidth}
>{\centering\arraybackslash}p{.073\textwidth}}
\toprule
\textbf{Method} & \textbf{SQA} & \textbf{MBPP} & \textbf{CoS-E} & \textbf{Tool} & \textbf{GPQA} & \textbf{HEval} & \textbf{Overall} & $\Delta$\\
\midrule
Raw             & 81.5 & 70.8 & 81.9 & 61.8 & 31.0 & 80.5 & 67.9 & \\
\midrule
\multicolumn{9}{l}{\emph{Baselines}} \\
SFT             & 80.8 & 70.4 & \textbf{82.6} & 66.2 & 30.6 & 79.3 & 68.3 & \resultgain{+0.4} \\
SDFT~\cite{shenfeld2026self}           & 81.6 & 71.6 & 81.2 & 73.5 & 34.2 & 78.7 & 70.1 & \resultgain{+2.2} \\
GKD~\cite{agarwal2024policy} & 81.2 & 72.8 & 81.8 & 72.1 & 33.0 & 82.3 & 70.5 & \resultgain{+2.6} \\
SSD~\cite{zhang2026embarrassingly} & 80.8 & 72.4 & 79.9 & 55.9 & 33.7 & 81.1 & 67.3 & \resultloss{-0.6} \\
OPSD~\cite{zhao2026self} & 81.2 & 72.8 & 82.0 & 61.8 & 31.5 & 79.9 & 68.2 & \resultgain{+0.3} \\
\midrule
\multicolumn{9}{l}{\emph{Variants of} \method} \\
Agree (Tok.) & \textbf{85.2} & 71.2 & \underline{82.2} & 75.0 & \underline{36.2} & \textbf{83.5} & 72.2 & \resultgain{+4.3} \\
Agree (Seq.) & 84.4 & 73.2 & 81.9 & \underline{76.5} & 35.7 & \textbf{83.5} & \underline{72.5} & \underline{\resultgain{+4.6}} \\
EMA             & 84.3 & 73.2 & 81.4 & \textbf{77.9} & 35.3 & \underline{82.9} & \underline{72.5} & \underline{\resultgain{+4.6}} \\
Contrast    & 83.9 & \underline{73.5} & 82.0 & 75.0 & 34.6 & 82.3 & 71.9 & \resultgain{+4.0} \\
Match (Joint) & 84.8 & 73.2 & 81.8 & \underline{76.5} & 33.5 & \underline{82.9} & 72.1 & \resultgain{+4.2} \\
Match (Repr.)  & 83.7 & 73.2 & 81.7 & 72.1 & 35.9 & 82.3 & 71.5 & \resultgain{+3.6} \\
Clip & 82.8 & 73.2 & 81.7 & 70.6 & 31.5 & 81.7 & 70.3 & \resultgain{+2.4}\\
\midrule
\textbf{\methodfull} & \underline{85.0} & \textbf{74.7} & \underline{82.2} & \textbf{77.9} & \textbf{36.4} & \textbf{83.5} & \textbf{73.3} & \textbf{\resultgain{+5.4}} \\
\bottomrule
\end{tabular}
\caption{
Results of \method variants, \methodfull, and baselines on ID and \shadetext{out-of-domain (in gray)} benchmarks using \texttt{Qwen2.5-7B} as the base model.
SQA, Tool, and HEval abbreviate ScienceQA, ToolAlpaca, and HumanEval, respectively.
%  In-domain evaluation includes ScienceQA, MBPP, CoS-E, and ToolAlpaca.
For agreement variants, we use retrieved contexts. Agree (Tok./Seq.) denotes token-/sequence-level agreement. EMA, Contrast, and Clip denote EMA teacher, token-level contrastive learning, and divergence clipping, respectively. Match (Repr./Joint) denotes representation-centric and joint logit--representation matching, respectively.
%: GPQA for advanced reasoning and HumanEval for coding capabilities.
$\Delta$ denotes change from Raw; green and red indicate increases and decreases, respectively.
\textbf{Bold} and \underline{underline} denote the best and second-best results, respectively.
% \yq{Give better names.}
}
\label{tab:overall_qwen25-7b}
\myvspace{-3mm}
\end{table*}

\subsection{Experimental Setup}
\label{sec:exp_setup}
\paragraph{Datasets and models.}
We evaluate six benchmarks across four task categories: scientific reasoning with \textsc{ScienceQA}~\cite{lu2022learn} and \textsc{GPQA}~\cite{rein2024gpqa}; commonsense reasoning with \textsc{CoS-E}~\cite{rajani2019explain}; code generation with \textsc{MBPP}~\cite{austin2021program} and \textsc{HumanEval}~\cite{chen2021evaluating}; and tool use with \textsc{ToolAlpaca}~\cite{tang2023toolalpaca}.
ScienceQA-trained models are tested on GPQA, and MBPP-trained models on HumanEval, for task-adjacent OOD transfer; the other four benchmarks provide training and in-domain evaluation.
Table~\ref{tab:dataset_stats} gives dataset statistics and licenses.
Our primary model is \texttt{Qwen2.5-7B-Instruct}~\cite{qwen25}.
We additionally evaluate Qwen2.5 at 0.5B, 1.5B, and 3B, \texttt{Llama-3.1-8B-Instruct}~\cite{dubey2024llama}, and \texttt{gemma-3-4b-it}~\cite{team2025gemma}. The implementation is based on PyTorch~\cite{paszke2019pytorch}, Transformers~\cite{wolf2020transformers}, and TRL~\cite{vonwerra2020trl} libraries. 
Appendix~\ref{app:exp_setup} specifies training, supervision construction, and greedy evaluation.
% The supplementary diagnostics use separately identified protocols.

\paragraph{Training and evaluation.}
Unless otherwise noted, we train the models with LoRA~\cite{hu2022lora} (rank 64, alpha 128, dropout 0.05) and AdamW~\cite{loshchilov2018decoupled} at $2\!\times\!10^{-5}$, using cosine decay and 10\% warmup.
On-policy generation uses temperature 0.7; evaluation is greedy.
Multiple-choice tasks use accuracy, code generation uses pass@1, and tool use requires exact matches of action names and all arguments.

\subsection{Main Results}
\label{sec:main}

\paragraph{The integrated recipe improves the six-benchmark mean.}
Table~\ref{tab:overall_qwen25-7b} compares Qwen2.5-7B baselines, individual controls, and \methodfull.
The integrated recipe reaches 73.3 overall, improving over Raw by 5.4 points and the strongest evaluated distillation baseline, GKD, by 2.8.
It is best or tied-best on MBPP, ToolAlpaca, GPQA, and HumanEval, and second-best on ScienceQA and CoS-E.
Appendix~\ref{app:score_uncertainty} reports pointwise confidence intervals for the two OOD benchmarks, conditional on the evaluated checkpoints.

\paragraph{Static imitation has task-dependent benefits.}
SFT improves ToolAlpaca by 4.4 points and CoS-E by 0.7; its ScienceQA, GPQA, MBPP, and HumanEval scores are below Raw.
On-policy training provides gains across a broader set of tasks in this comparison.
SDFT improves ToolAlpaca by 11.7 points and GPQA by 3.2, with HumanEval and CoS-E scores below Raw.
The training-split analysis in Appendix~\ref{app:train_split} further examines these differences.

\begin{figure*}[t]
\centering
\makebox[\widefigurewidth]{%
\includegraphics[width=0.38\widefigurewidth]{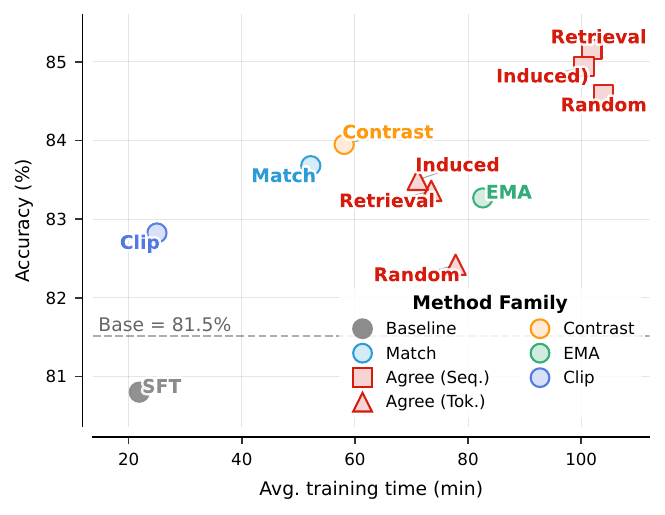}
\includegraphics[width=0.29\widefigurewidth]{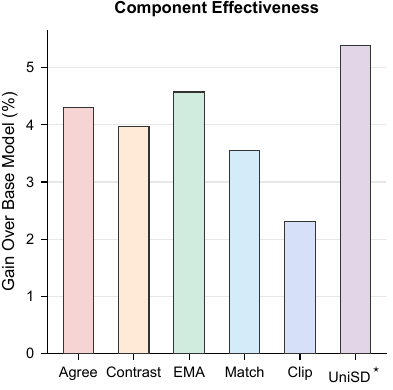}
\hfill
\includegraphics[width=0.29\widefigurewidth]{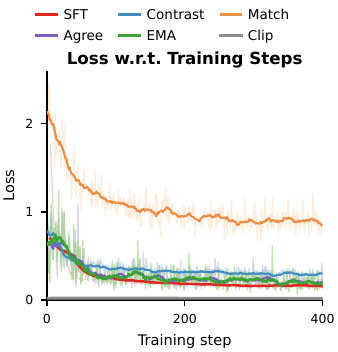}%
}
\myvspace{-2mm}
\caption{
\textbf{Left:} ScienceQA accuracy versus average training time.
\textbf{Middle:} Six-benchmark mean gains over Raw; bars show the plotted agreement and matching configurations.
\textbf{Right:} Training-loss trajectories on Qwen2.5-7B.
\methodfull has the largest mean gain ($+5.4$ points). Loss trajectories complement the train--test comparison in Appendix~\ref{app:train_split}.
}
\myvspace{-5mm}
\label{fig:training_time_component_loss}
\end{figure*}

\paragraph{Individual controls provide different benefits.}
Token-level agreement achieves the highest ScienceQA accuracy (85.2) and ties for the best HumanEval result.
The reported sequence-level configuration matches or improves Raw on every benchmark and attains a higher mean than token-level agreement (72.5 versus 72.2).
EMA also reaches 72.5 overall, with particularly large gains on ToolAlpaca (77.9, +16.1 over Raw).
In a separate single-seed update-rate sweep, increasing EMA retention from 0.50 to 0.99 improves ScienceQA from 80.4 to 82.6; the complete sweep and its OOD trade-off appear in Appendix~\ref{app:ema_retention}.
Contrastive learning improves all six benchmarks (71.9 overall).
Joint representation/logit matching reaches 72.1 versus 71.5 for representation-centric matching, supporting the use of both supervision spaces.
Clipping mainly provides a lightweight stabilizer (Figure~\ref{fig:resource_comparison}). 
Figure~\ref{fig:training_time_component_loss} summarizes the performance gains, training cost, and loss trajectories.

\paragraph{Conditional component contributions.}
We remove one component at a time under fixed settings (Appendix~\ref{app:loo}; Figure~\ref{fig:overview}a).
Removing agreement reduces accuracy by 3.5 points to 81.5; removing EMA, contrastive learning, or feature matching costs 2.5, 1.4, or 1.0 points, respectively.
Figure~\ref{fig:component_effectiveness} shows standalone effects by task.

\paragraph{The integrated recipe leads the longer-training comparison.}
Three-epoch SFT reaches 84.3/34.2/63.2 on ScienceQA/GPQA/ToolAlpaca, versus 85.0/36.4/77.9 for one-epoch \methodfull.
Three-epoch SDFT scores 82.0/31.5/69.1 (Appendix~\ref{app:compute_matched}; Figure~\ref{fig:compute_matched}).
This epoch-based control complements the separate resource analysis (Appendix~\ref{app:resource}) and larger-teacher comparison (Appendix~\ref{app:strong_teacher}).

\begin{figure*}[t]
\centering
\includegraphics[width=\widefigurewidth]{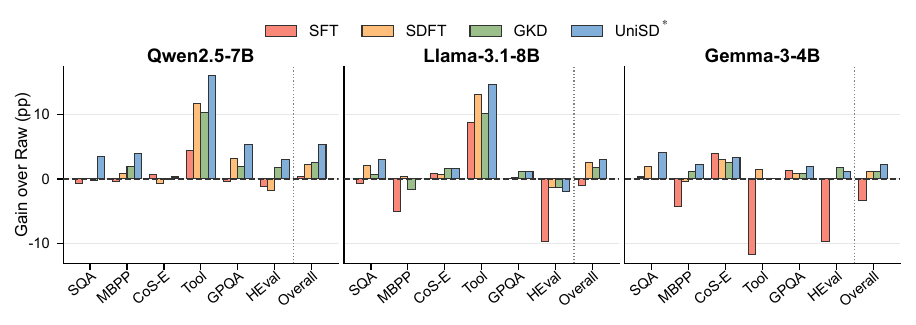}
% \myvspace{-6mm}
\caption{Gains over Raw across Qwen2.5-7B, Llama-3.1-8B, and Gemma-3-4B. Benchmark abbreviations follow Table~\ref{tab:overall_qwen25-7b}. \methodfull improves 15 of 18 model--dataset pairs.}
\label{fig:vary_model_family}
% \myvspace{-3mm}
\end{figure*}

\paragraph{EMA offers a lower-cost operating point.}
\methodfull improves or matches EMA on all six benchmarks, increasing the mean by 0.8 points.
Estimated energy is 0.26 versus 0.10 kWh per million training tokens ($+0.16$; Figure~\ref{fig:resource_comparison}).
Both deploy one adapted student without teacher scoring.

\subsection{Effects of Agreement Strategies}
\label{sec:agree}

\paragraph{Disagreement as a reliability signal.}
On 2{,}000 completions from 1{,}000 text-only ScienceQA prompts, incorrect completions have 2.2 times the mean sequence disagreement of correct ones (Appendix~\ref{app:disagreement_correctness}; Figure~\ref{fig:reliability}a).
Error rates are 16.2\% in the lowest-disagreement half and 36.5\% in the highest decile (panel b).
Sequence-level AUROC is 0.601 (95\% prompt-cluster CI [0.563, 0.637]; panel c), supporting disagreement as a graded signal of whole-completion reliability (Appendix~\ref{app:disagreement_correctness}).

\paragraph{Agreement strength and context sensitivity.}
Accuracy depends jointly on context count and agreement strength.
On ScienceQA, the token-level accuracy range across $K$ decreases from 2.20 points at $\gamma=0.01$ to 0.85 at $\gamma=1.0$.
Weaker penalties allow higher peaks, while stronger penalties yield less context-sensitive performance (Appendix~\ref{app:sensitivity}).
Varying the reliability normalization, cap, or floor changes accuracy by at most 0.8 points in the separate robustness check (Appendix~\ref{app:objective_spec}).

\paragraph{Context construction and granularity shape the gains.}
Figure~\ref{fig:radar_token_qwen25-7b} (Appendix~\ref{app:context_construction}) compares retrieval, random examples, and induced instructions.
Retrieval-based contexts provide nearest-neighbor examples and are most effective when semantic similarity offers task-specific evidence.
Under token-level agreement, retrieval gives the strongest results on ScienceQA (85.2), GPQA (36.2), and HumanEval (83.5).
The role of retrieval depends on the task: in coding and open-ended generation, nearest neighbors can share surface form while differing in valid implementation details, changing the information supplied by cross-context agreement.
Random contexts are more diverse and remain competitive across both token- and sequence-level agreement, suggesting that diverse, task-relevant examples can provide complementary supervision.
Induced contexts trade example-specific evidence for abstract task guidance. This particularly helps format-sensitive tasks: induced token-level agreement reaches 77.9 on ToolAlpaca, with smaller gains on CoS-E.
Under 5--20\% corrupted privileged information, Agree (Tok.) and \methodfull stay within 0.8 points of clean accuracy; SFT loses 5.2 points when corruption instead affects its training targets (Appendix~\ref{app:priv_noise}).

\paragraph{Distillation exceeds inference-time prompting.}
We test whether prompting can reproduce the gains by evaluating the raw model with retrieved examples, random examples, or induced instructions and no weight updates (Appendix~\ref{app:icl_control}).
The best prompted configurations reach 69.0 overall, versus 73.3 for the zero-shot \methodfull student.
On ToolAlpaca, the best prompted score is 67.6, compared with 77.9 for \methodfull.
The distilled student exceeds the best prompted result on five of six benchmarks while using no auxiliary context at inference.

\subsection{Generalization Across Models}
\label{sec:generalization_models}

\paragraph{Gains extend across the three tested families.}
Figure~\ref{fig:vary_model_family} compares task-level gains across families; six-benchmark mean gains are $+5.4$, $+3.1$, and $+2.2$ points for Qwen2.5, Llama-3.1, and Gemma-3.
Across 18 model--dataset pairs, \methodfull has 15 higher scores, two equal scores, and one lower score on an OOD task relative to Raw.
On Llama-3.1-8B, ScienceQA rises from 83.3 to 86.3 and ToolAlpaca from 57.4 to 72.1; HumanEval scores are 65.9 for Raw and 64.0 for \methodfull.
Appendix~\ref{app:cross_family_absolute} gives the complete absolute results.

\paragraph{The preferred control depends on model scale.}
Figure~\ref{fig:gain_and_retention_ppl_qwen25} compares model scales. EMA improves ScienceQA at both 0.5B and 1.5B, while agreement gains vary. \methodfull attains its largest ScienceQA gain at 3B (+7.06 points); GPQA rankings differ (Appendix~\ref{app:model_scale}; all gains are per task).

\paragraph{Task-adjacent transfer and far-OOD preservation differ.}
In the main comparison, \methodfull improves GPQA from 31.0 to 36.4 and HumanEval from 80.5 to 83.5.
Separate ScienceQA-to-MMLU-Pro and AIME 2024--2025 probes test more distant transfer.
On MMLU-Pro, \methodfull reaches 46.4 versus Raw's 47.6 and SFT's 41.4.
On AIME, it solves 5/60 problems versus Raw's 6/60.
These experiments measure capability retention under larger domain shifts (Appendix~\ref{app:far_ood}).

\myvspace{-1mm}
\section{Related Work}
\label{sec:related}
\myvspace{-1mm}
\paragraph{Continual Learning and On-Policy Learning.}
Continual learning~\cite{wang2024comprehensive} aims to adapt models to new knowledge and skills while preserving existing capabilities and mitigating catastrophic forgetting~\cite{mccloskey1989catastrophic,wang2026mascot}. 
In LLM post-training, this challenge is closely tied to the learning paradigm. 
Standard supervised fine-tuning (SFT) is off-policy, as it trains on fixed expert demonstrations rather than trajectories induced by the model's current policy, creating a training-inference mismatch. 
% a mismatch between training-time demonstrations and inference-time trajectories 
% as they train on fixed rather than trajectories induced by the model's current policy. 
On-policy learning reduces this mismatch by applying supervision to trajectories sampled from the current policy~\cite{song2026survey,agarwal2024policy,penaloza2026privileged}. 
% training on states the model actually visits during generation 
For example, GKD~\cite{agarwal2024policy} reduces exposure bias with on-policy sampling, while MiniLLM~\cite{guminillm} and DistiLLM~\cite{kodistillm} improve distribution matching through stabilized KL objectives. 
% As a result, continual adaptation based purely on offline imitation can damage prior capabilities, motivating methods that retain the benefits of on-policy learning. 

\paragraph{Knowledge Distillation and Self-Distillation for LLMs.}  
Knowledge distillation (KD)~\cite{hinton2015distilling} transfers knowledge from a teacher model to a student by matching predictions, logits, hidden states, generated outputs, or reasoning traces. 
Prior work distills token-level distributions, attention patterns, intermediate representations, rationales, and step-by-step reasoning traces from stronger models~\cite{jin2026entropy,luo2026agentark,hu2024visual}. 
Recent on-policy variants~\cite{zhong2026vla,zheng2026scope,luo2026demystifying} 
% such as VLA-OPD~\cite{zhong2026vla}, SCOPE~\cite{zheng2026scope}, and StableOPD~\cite{luo2026demystifying} 
further supervise student-generated trajectories using expert teachers or adaptive stabilization. 
% VLA-OPD uses an expert VLA teacher with a reverse-KL objective~\cite{zhong2026vla}, 
% while SCOPE and StableOPD improve OPD through adaptive weighting, reference constraints, and rollout mixture distillation~\cite{zheng2026scope,luo2026demystifying}. 
% VLA-OPD extends on-policy distillation to vision-language-action models by using an expert teacher to provide dense token-level supervision on the student’s self-generated trajectories under a reverse-KL objective. 
However, these methods usually depend on an external teacher model. 
Self-distillation instead derives supervision from the model itself or its variants, making it attractive when external teachers are costly, inaccessible, or undesirable~\cite{shenfeld2026self,zheng2026scope,hubotter2026reinforcement}. 
% This adaptation strategy is especially appealing when one has greater control over data, training, and deployment. 
% This distinction is orthogonal to on-policy learning: OPD specifies that the trajectories come from the student policy, whereas self-distillation specifies that the supervision is derived from the student or its variants rather than from an external teacher.
% \todo{Change to Self-Distillation (we work on SD but not OPD). Clearly distinguish between the two here: SD requires that the training signals are generated based on a stronger model, but OPD does not require that}
SDFT~\cite{shenfeld2026self} uses a demonstration-conditioned version of the base model as the teacher. OPSD~\cite{zhao2026self} distills dense supervision on student-generated trajectories.  
SDPO~\cite{hubotter2026reinforcement} uses privileged environment feedback for self-improvement. 
% SCOPE~\cite{zheng2026scope} introduces a dual-path adaptive framework that routes rollouts by correctness, applying teacher-perplexity-weighted KL distillation to incorrect trajectories and confidence-aware supervised learning to correct ones. 
However, most methods focus on individual strategies. \method explores the effectiveness of different mechanisms in a shared on-policy objective for controlled comparison.

\section{Conclusion}
\label{sec:conclusion}
\method enables reliability-aware on-policy self-distillation without stronger external teachers. Its controls offer task-dependent benefits; \methodfull achieves the highest six-benchmark mean in the primary Qwen2.5-7B comparison.

% Unnumbered main-text section required by ARR, placed after the conclusion.
\section*{Limitations}
\label{app:limitations}
Our experiments focus on established, bounded benchmarks spanning scientific and commonsense reasoning, code generation, and tool use. This scope enables controlled comparisons of self-distillation mechanisms across tasks and model families. Extending UniSD to interactive, long-horizon agentic settings, where feedback may be sparse or delayed, is a natural next step discussed in Appendix~\ref{app:future}. Our primary evaluation follows standard outcome-based protocols, including answer accuracy and executable correctness. 
Process-level evaluation could further characterize the effects on intermediate reasoning.
\section*{Ethical Considerations}
\label{app:ethics}
Responsible evaluation of self-distilled models should account for potential factual errors, social biases, and unsafe behaviors of the underlying base model.
Alongside \method's reliability weighting, divergence clipping, and stabilization, deployment should follow standard safety procedures.
Before deployment, adapted models should undergo safety, bias, factuality, and domain-specific risk evaluation, especially in human-centric settings.
% Users should obtain base models and benchmark datasets from their original providers and comply with the corresponding licenses, access restrictions, and use policies.

%++++++++++++++++++++++++++++++++++++++++
% References section will be created automatically 
% with inclusion of "thebibliography" environment
% as it shown below. See text starting with line
% \begin{thebibliography}{99}
% Note: with this approach it is YOUR responsibility to put them in order
% of appearance.

\clearpage
\bibliographystyle{unsrt}
\bibliography{reference}

\clearpage
\appendix

\newpage

\section{Algorithm Details of \method}
\label{sec-app-algo}

The detailed procedure of \method is shown in Algorithm~\ref{alg:method}.

\begin{algorithm*}[htbp]
\begin{algorithmic}[1]
\Require dataset $\mathcal D$, student $\pi_\theta$, training steps $N$, primary context $c$, auxiliary views $\{c^k\}_{k=1}^{K}$, enabled controls and coefficients
\State Initialize the teacher snapshot; set $\bar\theta\gets\theta$ if EMA is enabled
\For{$n=1,\ldots,N$}
    \State Sample $(x,c)\sim\mathcal D$ and a student completion $\hat y\sim\pi_\theta(\cdot\mid x)$
    \State Select the configured teacher snapshot or Exponential Moving Average (EMA) teacher $\pi^{\mathrm T}_{\bar\theta}$ under $c$
    \State Compute each token divergence $\mathcal D_t$: forward KL in the main configuration; reverse KL or Equation~\ref{eq:jsd} for the corresponding alternative
    \State Set $\widetilde{\mathcal D}_t\gets\min(\mathcal D_t,\kappa)$ if clipping is enabled; otherwise $\widetilde{\mathcal D}_t\gets\mathcal D_t$
    \If{agreement is enabled}
        \State Score the fixed completion under the $K$ auxiliary views (Equation~\ref{eq:aux_ll})
        \State Compute token- or sequence-level weights $w_t$ as specified in \S\ref{sec:framework}
    \Else
        \State Set $w_t\gets1$
    \EndIf
    \State $\displaystyle \mathcal L_{\mathrm{dist}}\gets\frac{\sum_t m_t w_t\widetilde{\mathcal D}_t}{\max(1,\sum_t m_t w_t)}$
    \State Initialize $\mathcal L_{\mathrm{aux}}\gets0$ and $\mathcal L_{\mathrm{feat}}\gets0$
    \If{contrastive learning is enabled}
        \State Score $\hat y$ under positive- and negative-conditioned teacher views
        \State Compute the normalized margin loss $\mathcal L_{\mathrm{aux}}$ (Equation~\ref{eq:contrast})
    \EndIf
    \If{feature matching is enabled}
        \State Compute normalized final-layer hidden-state MSE $\mathcal L_{\mathrm{feat}}$ (Equation~\ref{eq:feat})
    \EndIf
    \State $\mathcal L\gets\mathcal L_{\mathrm{dist}}+\lambda_{\mathrm{aux}}\mathcal L_{\mathrm{aux}}+\lambda_{\mathrm{feat}}\mathcal L_{\mathrm{feat}}$
    \If{the representation-only variant is selected}
        \State $\mathcal L\gets\mathcal L_{\mathrm{feat}}$
    \EndIf
    \State Update $\theta$ using the minibatch mean loss and the configured optimizer
    \If{EMA is enabled and $n$ is a synchronization step}
        \State $\bar\theta\gets\beta\bar\theta+(1-\beta)\theta$
    \EndIf
\EndFor
\end{algorithmic}
\caption{Training loop in per-completion form. Teacher scores and features are fixed targets; only student parameters receive gradients. The primary divergence and optional losses are selected per variant.}
\label{alg:method}
\end{algorithm*}

\subsection{Objective Specification of All Variants}
\label{app:objective_spec}

\begin{table}[t]
\centering
\footnotesize

\setlength{\tabcolsep}{3pt}
\begin{tabular}{@{}llll@{}}
\toprule
\textbf{Variant} & \textbf{Teacher} $\pi^{\mathrm T}_{*}$ & $w_t$ & $\mathcal L_{\mathrm{aux}}$ \\
\midrule
Agree (Tok.)  & Demo. & Tok.  & --- \\
Agree (Seq.)  & Demo. & Seq.  & --- \\
EMA           & EMA   & $1$   & --- \\
Contrast      & Demo. & $1$   & Eq.~\eqref{eq:contrast} \\
Match (Repr.) & Demo. & ---   & Eq.~\eqref{eq:feat} only \\
Match (Joint) & Demo. & $1$   & Eq.~\eqref{eq:feat}, hidden \\
Clip          & Demo. & $1$   & --- \\
\midrule
\methodfull   & EMA   & Tok.  & Eq.~\eqref{eq:contrast} + Eq.~\eqref{eq:feat} \\
\bottomrule
\end{tabular}
\caption{Objective specification of \method variants. All variants instantiate Equation~\ref{eq:overall} with matched data, rollout procedures, and optimization settings, and differ in the primary teacher $\pi^{\mathrm T}_{*}$, the reliability weight $w_t$, and the auxiliary objective $\mathcal L_{\mathrm{aux}}$. Match (Repr.) disables the primary divergence term; joint matching combines logit KL and hidden-state MSE. \emph{Demo.}\ denotes the demonstration-conditioned teacher; \emph{EMA} the EMA teacher $\pi^{\mathrm T}_{\bar\theta}$ (Eq.~\eqref{eq:ema}); \emph{Tok.}/\emph{Seq.}\ the token-/sequence-level agreement weight (Eq.~\eqref{eq:relweight}).}
\label{tab:objective_spec}
\end{table}

Every variant optimizes Equation~\ref{eq:overall} and differs in how the ingredients are instantiated: 1) the primary teacher $\pi^{\mathrm T}_{*}$, which is either the demonstration-conditioned teacher or the EMA teacher $\pi^{\mathrm T}_{\bar\theta}$ from Equation~\ref{eq:ema}; 2) the reliability weight $w_t$ from Equation~\ref{eq:relweight} (token-level), its sequence-level counterpart, or $w_t\equiv1$; 3) the divergence $D$, selected as forward KL in the main \method runs, with reverse KL and weighted JSD available as alternatives and optional clipping at $\kappa$; and 4) the auxiliary objective $\mathcal L_{\mathrm{aux}}$, which is the contrastive objective of Equation~\ref{eq:contrast}, the feature-matching term of Equation~\ref{eq:feat}, their sum, or absent ($\lambda_{\mathrm{aux}}=0$).
Table~\ref{tab:objective_spec} lists the variants. Match (Repr.) uses hidden-state MSE alone; Match (Joint) adds it to logit-distribution KL. Losses are normalized per completion and then averaged over the minibatch.
For the agreement variants, the auxiliary contextual teachers are used only to compute $w_t$; the teacher shown in the table provides the distillation target.
The final \methodfull objective is
\begin{equation}
\begin{split}
\mathcal L_{\mathrm{UniSD}^\ast}
&=\frac{\sum_t m_t w_t \widetilde{\mathcal D}_t}{\max(1,\sum_t m_t w_t)}\\
&\quad+\lambda_{\mathrm{aux}}\mathcal L_{\mathrm{aux}}
+\lambda_{\mathrm{feat}}\mathcal L_{\mathrm{feat}}.
\end{split}
\end{equation}
where $w_t$ is the token-level agreement weight, $\widetilde{\mathcal D}_t$ the divergence of Equation~\ref{eq:clip_obj} toward the EMA teacher $\pi^{\mathrm T}_{\bar\theta}$, $\mathcal L_{\mathrm{aux}}$ the contrastive objective of Equation~\ref{eq:contrast} with margin $\gamma_{\mathrm m}=0.1$, and $\mathcal L_{\mathrm{feat}}$ the hidden-state MSE of Equation~\ref{eq:feat}; logit matching is supplied by the primary KL term.
We set the corresponding loss coefficients to $\lambda_{\mathrm{aux}}=\lambda_{\mathrm{feat}}=0.01$.
The remaining hyperparameters are retrieved contexts with $K{=}3$, agreement strength $\gamma{=}0.001$, and the EMA update of Equation~\ref{eq:ema} (Appendix~\ref{app:exp_setup}).

\begin{figure}[t]
\centering
\includegraphics[width=\singlefigurewidth]{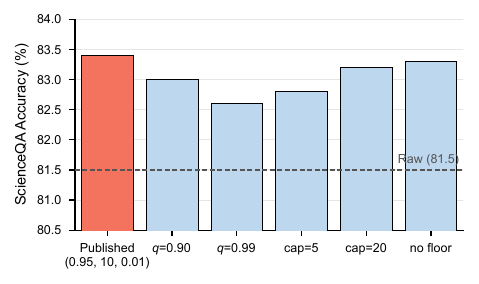}
\caption{Sensitivity to the reliability-weight constants on ScienceQA: Agree (Tok.), retrieved contexts, $K{=}3$, $\gamma{=}0.01$. Varying one constant of Equation~\ref{eq:relweight} at a time changes accuracy by at most 0.8 points; all settings exceed Raw (dashed line).}
\label{fig:constants_robustness}
\end{figure}

\paragraph{Rationale for the reliability-weight constants.}
The three constants in Equation~\ref{eq:relweight} are fixed robustness safeguards in the main comparisons. We vary them only in the separate sensitivity check below. Agreement-strength sweeps appear in Figures~\ref{fig:sensitivity_sequence_qwen25-7b}--\ref{fig:sensitivity_token_qwen25-7B}.
\emph{Quantile normalization} ($q_{0.95}$) provides scale calibration: the raw variance of teacher log-probabilities has a task- and batch-dependent scale, so dividing by a robust upper quantile makes $\hat{\delta}_t$ dimensionless and allows a single $\gamma$ to transfer across tasks; we use a high quantile rather than the batch maximum because the maximum is determined by exactly the rare outlier tokens the mechanism is designed to down-weight.
\emph{The cap at 10} bounds the leverage of extreme outliers beyond the normalization point, following the same robustness principle as divergence clipping, and guarantees $w_t \ge 1/(1+10\gamma)$ so reliability weighting remains positive.
\emph{The floor at 0.01} keeps reliability weights positive; for all $\gamma \le 1$ used in our experiments the cap already implies $w_t \ge 0.09 > 0.01$, so the floor is never active in practice and serves only as a numerical safeguard.
Figure~\ref{fig:constants_robustness} reports limited sensitivity over the tested ranges.
This separate Agree (Tok.) comparison is conducted on ScienceQA with retrieved contexts, $K{=}3$, and $\gamma{=}0.01$, varying one constant at a time: the normalization quantile over $q\in\{0.90,0.95,0.99\}$, the cap over $\{5,10,20\}$, and the floor over $\{0,0.01\}$.
The reference setting scores 83.4 in this separate diagnostic. Accuracy changes by at most 0.8 points across these settings, so the published $q_{0.95}$ choice is robust within the tested $0.90$--$0.99$ interval.

% Data (as a reference)
% Setting          q     cap  floor  ScienceQA
% Published        0.95  10   0.01   83.4
% Lower quantile   0.90  10   0.01   83.0
% Higher quantile  0.99  10   0.01   82.6
% Smaller cap      0.95   5   0.01   82.8
% Larger cap       0.95  20   0.01   83.2
% No floor         0.95  10   0      83.3

\section{Additional Experiments}

\subsection{Disagreement Diagnostic Protocol}
\label{app:disagreement_correctness}

\begin{figure*}[t]
\centering
\includegraphics[width=\widefigurewidth]{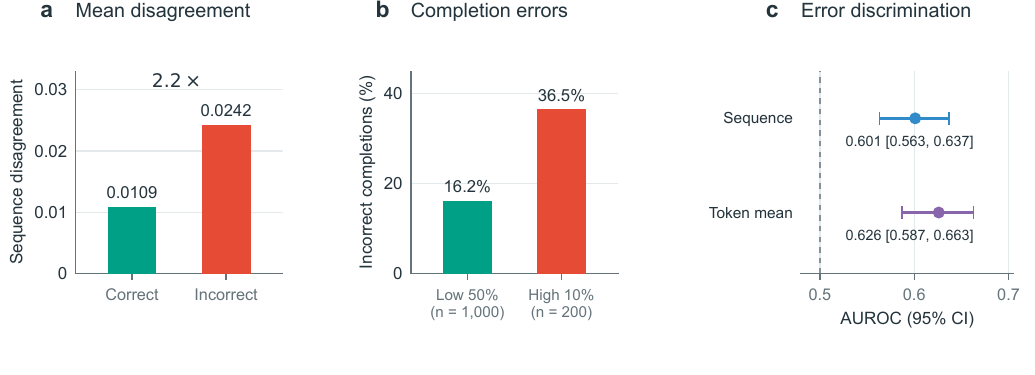}
\caption{Disagreement and completion correctness on 1{,}000 text-only ScienceQA prompts (two completions each).
\textbf{a,} Mean sequence disagreement.
\textbf{b,} Error rates in the lowest-disagreement half and highest-disagreement decile.
\textbf{c,} Completion-level AUROC with 95\% prompt-cluster bootstrap intervals; dashed line: chance (0.5).}
\label{fig:reliability}
\end{figure*}

Figure~\ref{fig:reliability} tests whether cross-view disagreement is associated with completion errors, independently of downstream accuracy.
The raw Qwen2.5-7B-Instruct model generates two completions for each of 1{,}000 text-only ScienceQA training prompts at temperature 0.7, with a maximum completion length of 1{,}024 tokens.
Each fixed completion is scored under three retrieved auxiliary teacher views using the disagreement definitions in \S\ref{sec:framework}.
Of the 2{,}000 completions, 417 are incorrect (20.85\%).
Correctness labels are used only for this diagnostic, not to compute training weights.

We quantify uncertainty with 10{,}000 bootstrap resamples of the 1{,}000 prompts, retaining both completions from each sampled prompt together.
This accounts for within-prompt dependence.
Both AUROCs use whole-completion correctness labels; the annotation and evaluation unit is the complete response.

\subsection{Sensitivity to EMA Retention}
\label{app:ema_retention}

We isolate teacher update rate in a separate, single-seed Qwen2.5-7B experiment.
Each run trains for one epoch on the same 2{,}000 text-only ScienceQA examples, while agreement, contrastive learning, feature matching, and clipping are disabled.
The teacher is synchronized every ten optimizer steps using the retention convention of Equation~\ref{eq:ema}.
All other settings are held fixed, and $\beta=1$ freezes the initial teacher.
Greedy evaluation uses 2{,}224 text-only ScienceQA test items and 448 GPQA questions.
This diagnostic uses the dedicated training and evaluation protocol specified here.

\begin{figure*}[t]
\centering
\includegraphics[width=\widefigurewidth]{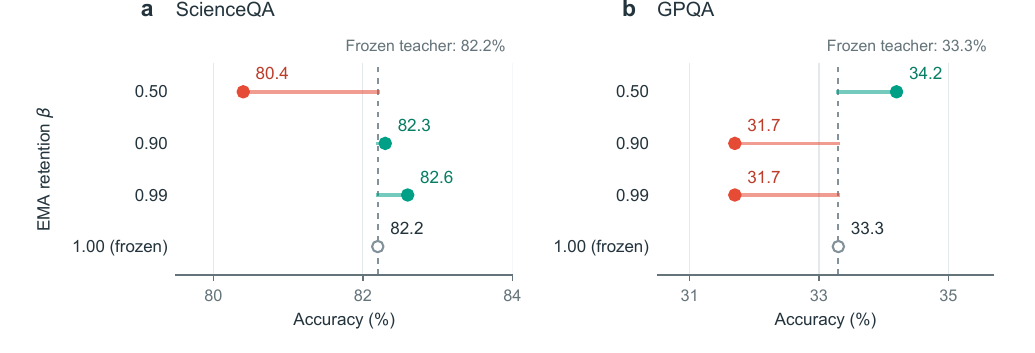}
\caption{Complete EMA-retention sweep after training on 2{,}000 text-only ScienceQA examples.
\textbf{a,} ScienceQA accuracy on 2{,}224 test items.
\textbf{b,} GPQA accuracy on 448 questions.
Dashed lines mark the frozen teacher; green/red encode the sign of the accuracy difference from this reference. Panels use separate accuracy scales, and every point is labeled with its score.}
\label{fig:ema}
\end{figure*}

Increasing retention from 0.50 to 0.99 improves ScienceQA by 2.2 points, supporting slower teacher updates in this setting.
The best setting exceeds the frozen teacher by 0.3 points, while GPQA favors a different retention setting.
The sweep characterizes how teacher retention affects in-domain accuracy and OOD transfer, with downstream accuracy as the measured outcome.

\subsection{Finite-Test-Set Uncertainty}
\label{app:score_uncertainty}

\methodfull scores 36.4\% on GPQA (163/448; 95\% Wilson CI: 32.1--40.9\%) and 83.5\% on HumanEval (137/164; 95\% Wilson CI: 77.1--88.4\%).
These pointwise intervals quantify finite-test-item uncertainty for each evaluated checkpoint under fixed decoding. The sampling unit is the test item.

\subsection{Training Time}
\label{app:time}

\begin{figure*}
\centering
\includegraphics[width=0.48\widefigurewidth]{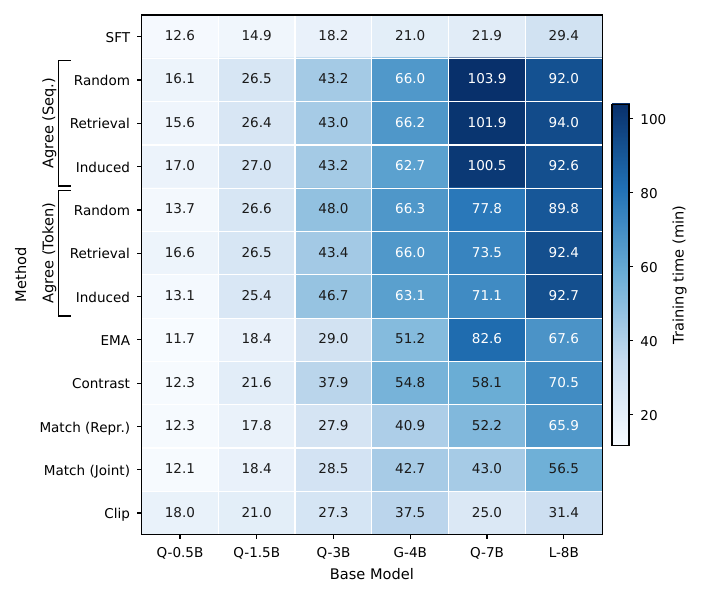}
\includegraphics[width=0.48\widefigurewidth]{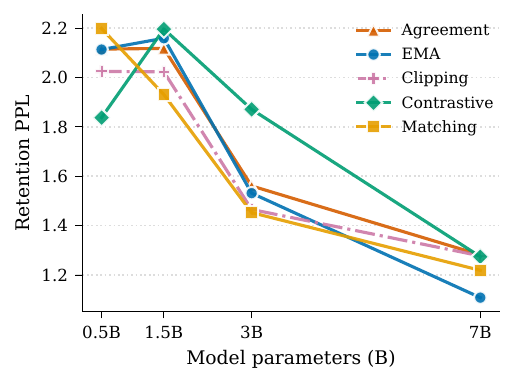}
\myvspace{-3mm}
\caption{
\textbf{Left.} Training time comparison of UniSD variants on ScienceQA. \textbf{Right.} Retention perplexity comparison across UniSD variants.
}
\label{fig:heatmap_training_time}
\end{figure*}

\paragraph{Training efficiency.}
Figure~\ref{fig:heatmap_training_time} compares the wall-clock training cost of different \method variants. 
For the \emph{agreement} setting, the main cost driver is not the distillation loss itself, but the number of teacher-conditioned scoring passes required for each on-policy completion. 
Standard SFT is the cheapest baseline. 
Agreement-based methods score each sampled completion under multiple auxiliary contexts before computing reliability weights.
For example, on Qwen2.5-7B, sequence-level agreement takes about $100$ minutes, compared with $21.9$ minutes for SFT. 
This connects the training cost of agreement estimation to the number of reliability views.

The comparison also reveals a useful design trade-off. 
Methods that add lightweight stabilization on top of a single teacher signal, such as clipping or feature matching, incur much smaller overhead than full multi-context agreement. 
EMA, contrastive learning, and joint matching lie between these extremes because they require additional teacher or auxiliary forward passes, but do not multiply the context-conditioned scoring as aggressively as agreement-based variants. 
Thus, future self-distillation systems should treat reliability estimation as a budgeted component: expensive multi-view agreement can be reserved for noisy or high-uncertainty examples, while cheaper stabilizers such as clipping, EMA smoothing, or representation matching can be applied broadly. 
This points to adaptive self-distillation designs that allocate computation according to signal reliability rather than applying the most expensive mechanism uniformly to every example.

\subsection{Resource Consumption}
\label{app:resource}
We evaluate the practical trade-offs of LLM post-training through task accuracy and resource consumption.
Prior work has emphasized that training cost affects not only environmental impact, but also reproducibility and accessibility for researchers with limited compute~\cite{strubell2019energy,schwartz2020green,patterson2021carbon}.
We therefore complement the wall-clock analysis in \App~\ref{app:time} with estimated resource consumption. 
Since absolute runtime depends on batching, memory budget, and hyperparameters, we report token-normalized cost: energy per million training tokens and throughput in million tokens per GPU-hour.
These metrics quantify on-policy generation and scoring costs.

% Following previous emissions accounting methods in energy-aware ML
Following the emissions accounting used by CodeCarbon~\cite{benoit_courty_2024_11171501} and the MLCO$_2$ Impact Calculator~\cite{lacoste2019quantifying}, we first estimate energy consumption from runtime and then convert it to $\mathrm{CO}_2$-equivalent emissions using grid carbon intensity.
For each completed training run, we estimate energy from runtime under fixed hardware and utilization assumptions:
\begin{equation}
\mathrm{kWh}
=
T \cdot N_{\mathrm{GPU}} \cdot
    \frac{P_{\mathrm{TDP}}}{1000}
\cdot u \cdot \mathrm{PUE},
\end{equation}
where $T$ is wall-clock time in hours, $N_{\mathrm{GPU}}$ is the number of GPUs, $P_{\mathrm{TDP}}=300\mathrm{W}$ is the TDP of an NVIDIA A100 PCIe 80GB GPU, $u=0.7$ is the assumed sustained utilization, and $\mathrm{PUE}=1.2$ is Power Usage Effectiveness. 
PUE is the ratio between total data-center energy and IT equipment energy, accounting for facility overhead such as cooling and power delivery losses~\cite{avelar2012pue}. 
We then estimate emissions as
\begin{equation}
\mathrm{kgCO_2e}
=
\mathrm{kWh} \cdot \frac{c}{1000},
\end{equation}
where $c=475\mathrm{gCO_2e/kWh}$ is the assumed carbon intensity. 
The resulting model-based estimates support relative comparisons under these shared assumptions.

\begin{figure*}[t]
\centering
\includegraphics[width=\widefigurewidth]{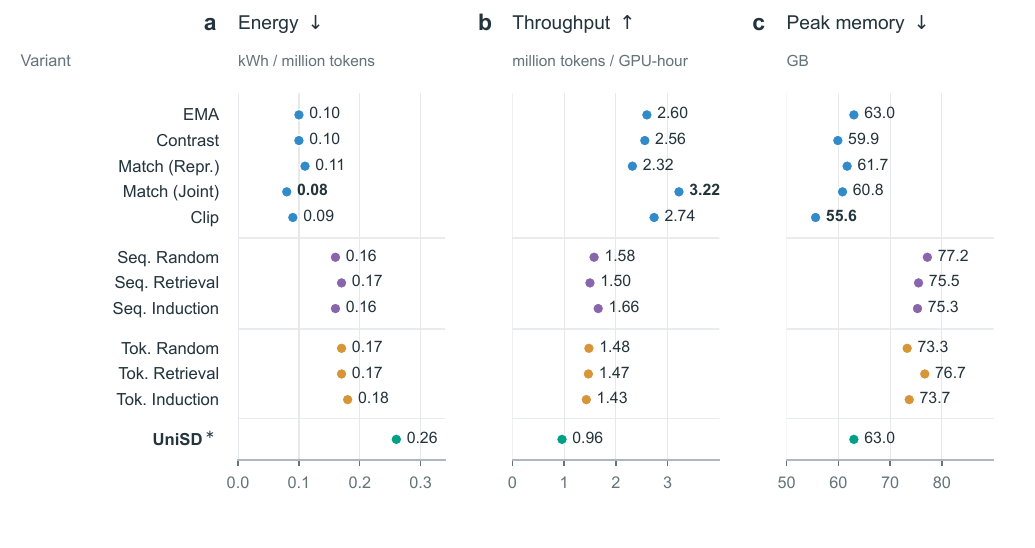}
\caption{Estimated resource consumption of \method variants:
\textbf{a,} energy per million training tokens;
\textbf{b,} throughput in million tokens per GPU-hour;
\textbf{c,} peak memory.
Every point is labeled with its reported value; bold values mark the best value for each metric.
Seq.\ and Tok.\ denote sequence- and token-level agreement.
Energy estimates use NVIDIA A100 PCIe 80GB TDP ($300$W), utilization $u=0.7$, and $\mathrm{PUE}=1.2$; the emissions calculation assumes $475\,\mathrm{gCO_2e/kWh}$.
These runtime-based estimates support relative comparison under the stated assumptions.}
\label{fig:resource_comparison}
\end{figure*}

\paragraph{Token-normalized cost and memory footprint.}
Figure~\ref{fig:resource_comparison} compares the token-normalized cost of \method variants. 
% SFT is the cheapest, requiring 0.05 kWh per million tokens. 
Single-teacher stabilization methods are more efficient. Match (Joint) requires only 0.08 kWh per million tokens, while Contrast, EMA, and Match (Repr.) require 0.10--0.11 kWh per million tokens. 
These variants preserve high throughput (2.32--3.22M tokens/GPU-hour), showing that adding representation, contrastive, or temporal stabilization incurs only modest overhead. 
Agreement-based variants require 0.16--0.18 kWh per million tokens and use 73.3--77.2GB of peak memory, 13.1--17.0GB above the five single-teacher variants' mean of 60.2GB (+22--28\%). 
This overhead is expected: Agreement estimates reliability by re-scoring each on-policy completion under multiple auxiliary contexts, increasing teacher-side forward computation and storing additional prompt--completion tensors, masks, and log-probability buffers.
With multi-context scoring, throughput is 1.43--1.66M tokens/GPU-hour; the additional computation and memory provide a consistency signal for weighting self-supervision.
% The integrated \methodfull further increases token-normalized energy to 0.26 kWh per million tokens and reduces throughput to 0.96M tokens/GPU-hour, reflecting the cost of combining multiple reliability mechanisms in one pipeline. 
Implementations with tighter memory budgets can reduce Agreement overhead by scoring auxiliary contexts sequentially rather than jointly. 

\begin{figure*}[t]
\centering
\includegraphics[width=\widefigurewidth]{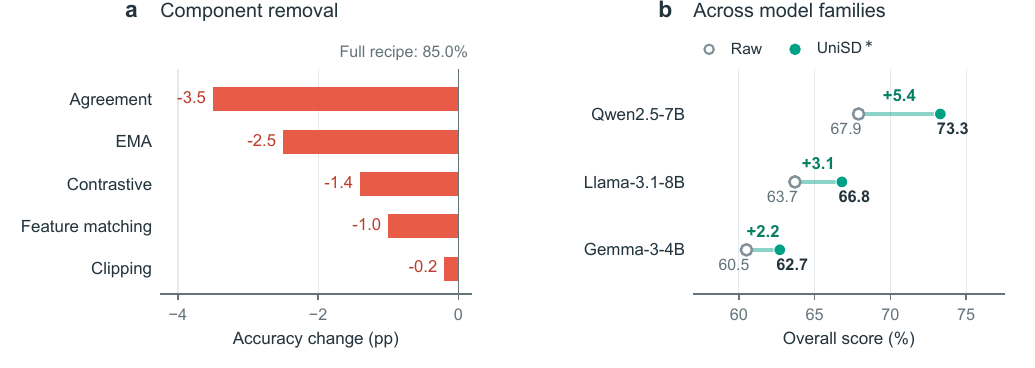}
\caption{Conditional contributions and cross-family gains.
\textbf{a,} Leave-one-out accuracy changes from \methodfull (85.0\%) under the 2{,}000-example ScienceQA protocol.
\textbf{b,} Six-benchmark means for Raw and \methodfull across three model families; annotations show gains in percentage points.}
\label{fig:overview}
\end{figure*}

Figure~\ref{fig:overview} summarizes performance; Appendices~\ref{app:loo} and \ref{app:cross_family_absolute} give the component-removal protocol and absolute cross-family results.

\subsection{Completion Likelihood and Distribution Retention}
\label{sec:distribution}
\myvspace{-1mm}
% Alternative Titles:
% Likelihood and Retention Analysis
% Likelihood and Distributional Retention Analysis (Completion Likelihood)
% Likelihood, Retention, and Distributional Drift
% Gold-Completion Likelihood and Retention
To characterize how adaptation changes the model distribution, we evaluate two properties complementary to task accuracy.
First, \emph{reference-completion fit} % tests whether the adapted model assigns higher likelihood to the gold answer trajectory
goes beyond final-answer accuracy and measures whether the adapted model makes the gold completion more likely under teacher forcing. % This goes beyond final-answer accuracy by checking whether the model assigns higher probability to the full reference. 
Second, \emph{distributional retention} measures whether the adapted model preserves the base model's original generative behavior while acquiring the target skill~\cite{yang2024federated}. 
Base-scored likelihood characterizes distributional compatibility. Complementary downstream evaluations assess capability retention after adaptation, the concern underlying catastrophic forgetting~\cite{mccloskey1989catastrophic} (Appendix~\ref{app:far_ood}).

\begin{figure*}[t]
\centering
\includegraphics[width=\widefigurewidth]{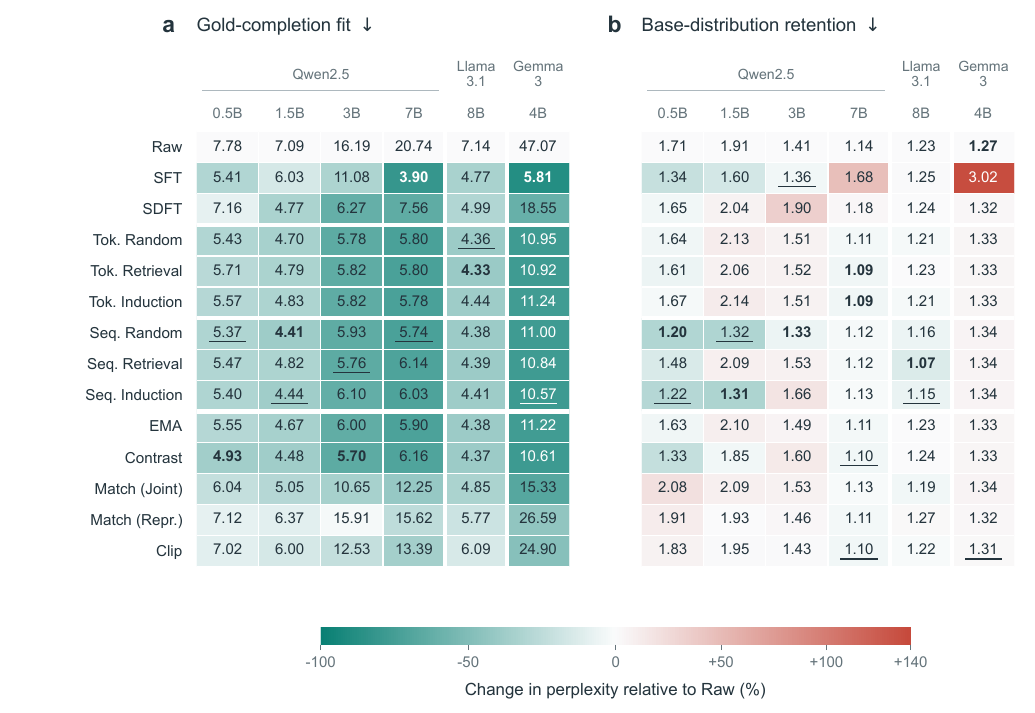}
\caption{Complementary perplexity diagnostics across model families:
\textbf{a,} teacher-forced perplexity on gold completions;
\textbf{b,} base-distribution retention perplexity (\S\ref{sec:distribution}).
Cell labels preserve all reported absolute values.
Color shows the percentage change from Raw within each model and metric on a shared scale: green indicates lower perplexity and red higher perplexity.
Bold and underlined values mark the lowest and second-lowest distinct perplexities in each column, including ties. 
Tok.\ and Seq.\ denote token- and sequence-level agreement.}
\label{fig:perplexity_comparison}
\end{figure*}

\paragraph{Gold-completion fit.}
Given a prompt--completion pair $(x, y)$, we measure whether adaptation improves the likelihood of the gold completion by scoring completion tokens under teacher forcing: 
\begin{equation*}
\mathrm{PPL}_{\mathrm{fit}}
=
\exp\!\left(
-\frac{
\sum\limits_{\substack{(x,y)\in\mathcal D\\t\in\mathcal M(x,y)}}
\log p_\theta(y_t \mid x,y_{<t})
}{
\sum\limits_{(x,y)\in\mathcal D} |\mathcal M(x,y)|
}
\right),
\end{equation*}
where $\mathcal M(x,y)$ denotes completion-token positions. 
By scoring only completion tokens, this metric focuses on how well the model supports the desired answer trajectory. 
Across model families, self-distillation substantially improves reference-completion likelihood (Figure~\ref{fig:perplexity_comparison}a). 
On Qwen2.5-7B, Agreement variants, EMA, and Contrast reduce perplexity from 20.74 to 5.74--6.16. On Gemma-3-4B, these variants reduce perplexity from 47.07 to 10.57--11.24.
SFT, which directly optimizes gold completions, reaches 3.90 and 5.81 on these two models; each of the six SFT fit evaluations scores all 2,224 ScienceQA test examples.
% These gains suggest that reliable token-level supervision is central to effective self-distillation.
Feature matching has model-dependent effects on completion likelihood, consistent with its role as an auxiliary regularizer.
% rather than a major supervision signal. 
% Future works can focus on verifying supervision through agreement across contextual views. 

\paragraph{Base-distribution retention.} 
% Beyond task accuracy, effective adaptation should manifest \emph{retention}~\cite{yang2024federated}, i.e. preserve the base model's original generative behavior while acquiring target skill. 
% We define this property as \emph{retention}: the ability to acquire the target skill without excessive drift from the base distribution. 
% Poor retention reflects a form of catastrophic forgetting, where task-specific updates overwrite previously acquired capabilities~\cite{mccloskey1989catastrophic}. 
% Such models may appear successful on the adaptation task but become over-specialized, producing generations that are less natural under the base model. 
We next measure \emph{distributional retention}, i.e.  whether adapted generations remain likely under the original base model. 
For each prompt $x$, we sample a completion $\hat{y}^{x}$ from the adapted model and score it under the original base model $\pi_0$:
\begin{equation*}
\mathrm{PPL}_{\mathrm{ret}}
=
\exp\!\left(
-\frac{
\sum\limits_{x\in\mathcal D}\sum\limits_{t}
\log \pi_0(\hat{y}_t^{x}\mid x,\hat{y}_{<t}^{x})
}{
\sum\limits_{x\in\mathcal D} |\hat{y}^{x}|
}
\right).
\end{equation*}
Lower $\mathrm{PPL}_{\mathrm{ret}}$ indicates that adapted generations remain more likely under the base distribution, complementing reference-completion fit by measuring preservation rather than task fit. 
Figure~\ref{fig:perplexity_comparison}b shows that SFT can induce substantial drift: on Qwen2.5-7B, retention perplexity increases from 1.14 for the raw model to 1.68, while on Gemma-3-4B it rises from 1.27 to 3.02. 
On Qwen2.5-7B and Gemma-3-4B, the on-policy variants avoid SFT's sharp retention-perplexity increases.
For Qwen2.5-7B, Agreement, EMA, Contrast, and Clip keep $\mathrm{PPL}_{\mathrm{ret}}$ close to the raw model, with the best values between 1.09 and 1.13. 
% Within Qwen2.5-7B, the strongest retention improvement comes from stabilizing the teacher: 
EMA teacher reduces retention perplexity by 33.9\% relative to SFT, suggesting that a smoothly evolving teacher provides a more distribution-compatible target. 

% Regular float (not wrapfigure): wrapfig collapses the text column in the
% two-column ACL build, and the appendix has no page limit in either venue.
% Fixed width renders the same physical size in one- and two-column layouts.
\begin{figure*}[t]
\centering
\includegraphics[width=\widefigurewidth]{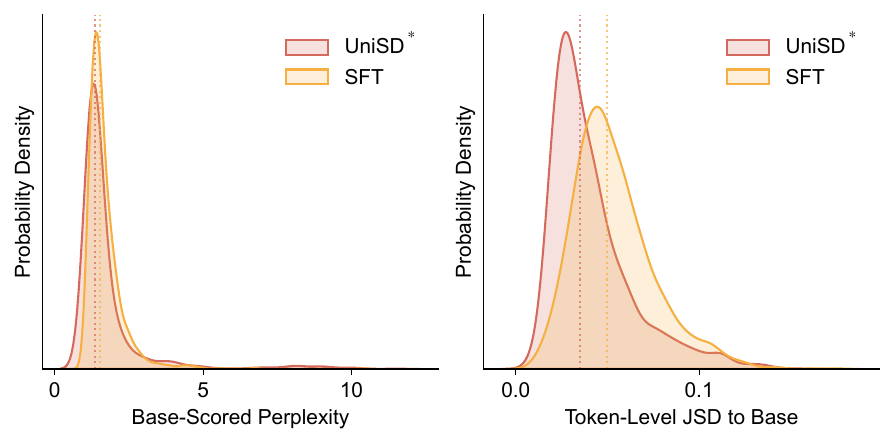}
\myvspace{-2mm}
\caption{Distribution of base-scored perplexity and token-level Jensen--Shannon divergence (JSD).}
\label{fig:distribution}
\myvspace{-3mm}
\end{figure*}

Figure~\ref{fig:distribution} further examines retention at the trajectory level. 
For each generated completion, we compute both base-scored perplexity and the average token-level JSD between the adapted and base next-token distributions along the same trajectory. 
\methodfull improves accuracy from 80.8 to 85.0 while reducing mean token-level JSD from 0.054 for SFT to 0.041. 
The paired analysis further shows that \methodfull has lower JSD than SFT on 70.3\% of examples, with both the mean and median paired differences below zero. 
Similarly, the base-log-probability comparison shows that \methodfull completions receive higher base-model log-probability on 60.6\% of examples. 
% The paired analysis as shown in further strengthens this conclusion. On matched examples, our method yields lower JSD than SFT on 70.3 of prompts, with both the mean and median paired differences below zero, showing that the retention advantage is broad and not driven by a few outliers. The log-probability comparison points in the same direction but is weaker: completions from our method receive higher base-model log-probability on 60.6 of examples. Overall, these results suggest that JSD is the more sensitive retention metric here, since it captures full next-token distribution shift rather than only the likelihood of the sampled sequence.
Together, these trajectory-level diagnostics show greater compatibility with the base model in both generated completions and token-level predictive distributions.

\begin{figure*}[t]
\centering
\includegraphics[width=0.9\widefigurewidth]{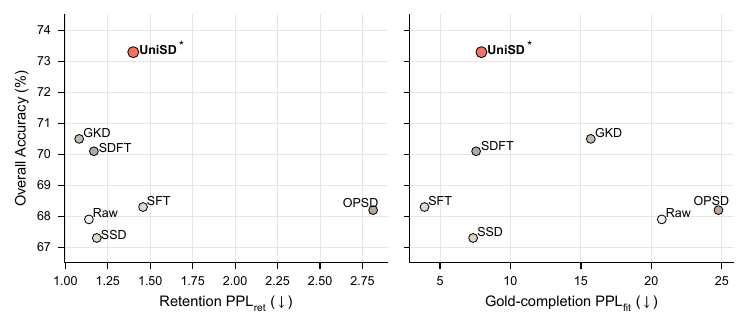}
\caption{Method-level summary for Qwen2.5-7B: six-benchmark means from Table~\ref{tab:overall_qwen25-7b} versus reported ScienceQA-checkpoint perplexity. \methodfull has the highest mean and lower retention perplexity than SFT (1.40 vs.\ 1.46). The axes summarize distinct evaluation scopes at the method level.}
\label{fig:retention_pareto}
\end{figure*}

\paragraph{Method-level accuracy and retention summary.}
Figure~\ref{fig:retention_pareto} provides a descriptive method-level comparison: one axis reports six-benchmark means from Table~\ref{tab:overall_qwen25-7b}, and the other reports ScienceQA-checkpoint perplexity from a separate evaluation export. Each axis retains its own evaluation scope. The retention export contains aggregate method-level perplexities without retained-item counts.
GKD achieves the lowest $\mathrm{PPL}_{\mathrm{ret}}$ and a six-task mean of 70.5 vs.\ 73.3 for \methodfull; SSD pairs base-distribution retention with overall accuracy at or below Raw, while OPSD has higher retention perplexity.
SFT has the lowest reported $\mathrm{PPL}_{\mathrm{fit}}$, consistent with its direct gold-completion objective, and a six-task accuracy change from 67.9 to 68.3. Reference fit and test accuracy capture distinct aspects of adaptation; Appendix~\ref{app:train_split} examines the train--test patterns.
\methodfull has the highest six-task mean and lower reported retention perplexity than SFT, while GKD has lower retention perplexity. The plot summarizes these complementary method-level outcomes.

% Data (as a reference)
% Method  Overall Acc.  PPL_ret  PPL_fit
% Raw     67.9          1.14     20.74
% SFT     68.3          1.46      3.90
% SDFT    70.1          1.17      7.56
% GKD     70.5          1.08     15.70
% SSD     67.3          1.19      7.35
% OPSD    68.2          2.81     24.76
% UniSD*  73.3          1.40      7.94

% Absolute metrics alone do not capture how much an adapted model drifts from its original distribution.
% We additionally measure whether each method's generations remain likely under the corresponding raw model, ensuring that each method improves task performance while preserving prior capabilities.
% Intuitively, if the raw model still assigns high likelihood to an adapted model's outputs, the adaptation induces less distributional drift.
% If not, the model may have shifted toward a narrower or less compatible generation style.

\subsection{Comparison with Stronger External Teachers}
\label{app:strong_teacher}

\begin{figure*}[t]
\centering
\includegraphics[width=0.8\widefigurewidth]{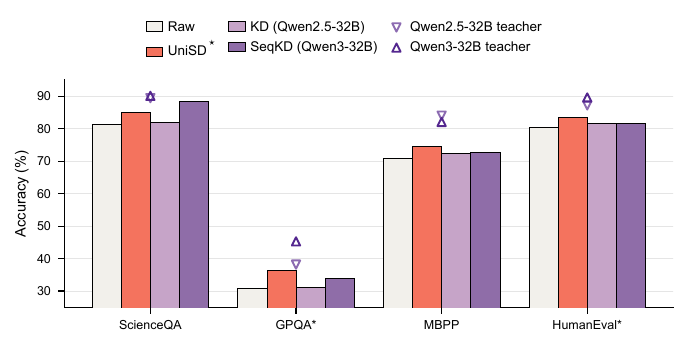}
\caption{Distillation from external teachers ${\sim}4.5\times$ the student's size. Bars compare the raw student, \methodfull (no external model), token-level KD from Qwen2.5-32B, and sequence-level KD from Qwen3-32B; open markers show the 32B teachers' zero-shot scores. Asterisks denote OOD benchmarks. \methodfull outperforms token-level KD from the 32B teacher on all four benchmarks.}
\label{fig:strong_teacher}
\end{figure*}

We measure the gap to distillation from external teachers roughly $4.5\times$ the student's size in two settings (Figure~\ref{fig:strong_teacher}).
First, we evaluate token-level KD from Qwen2.5-32B-Instruct~\cite{qwen25}, using the larger model to score student-generated trajectories.
Second, we run sequence-level KD from Qwen3-32B~\cite{yang2025qwen3}, using its generated responses to the same training prompts.
The teachers retain a substantial zero-shot advantage over the raw student (7.9--8.7 points on ScienceQA and 7.2--14.3 on GPQA), yet distillation transfers only part of it.
\methodfull outperforms the reported token-level KD configuration on all four benchmarks (85.0/36.4/74.7/83.5 vs.\ 82.1/31.2/72.4/81.7). This comparison highlights the value of the student-training recipe alongside the teacher's zero-shot capability.
Sequence-level KD from Qwen3-32B recovers most of the in-domain ScienceQA gain (88.5) but transfers less effectively out of domain, underperforming \methodfull on GPQA, MBPP, and HumanEval.
The two paradigms are therefore complementary: strong-teacher KD requires access to and inference from a substantially larger external model, whereas \method targets settings where such teachers are costly, inaccessible, or undesirable (\S\ref{sec:intro}).

% Data (as a reference)
% Method                     ScienceQA  GPQA  MBPP  HumanEval
% Raw Qwen2.5-7B-Instruct    81.5       31.0  70.8  80.5
% UniSD* (ours)              85.0       36.4  74.7  83.5
% Token-level KD Qwen2.5-32B 82.1       31.2  72.4  81.7
% SeqKD Qwen3-32B            88.5       33.9  72.8  81.7
% Qwen2.5-32B teacher        89.4       38.2  84.0  87.2
% Qwen3-32B teacher          90.2       45.3  82.1  89.6

\subsection{Prompted-Teacher Control}
\label{app:icl_control}

\begin{figure*}[t]
\centering
\includegraphics[width=0.95\widefigurewidth]{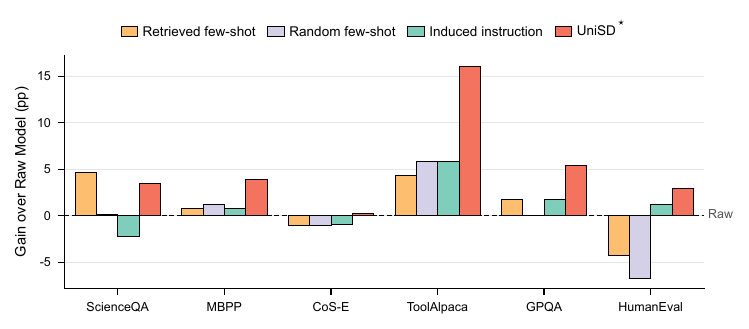}
\myvspace{-2mm}
\caption{Prompted-teacher control. Gains over Raw from inference-time prompting with the training context strategies, compared with the zero-shot \methodfull student. Few-shot HumanEval and CoS-E scores are below Raw; \methodfull leads the prompted configurations on ToolAlpaca. The zero-shot student exceeds the best prompted result on five of six benchmarks.}
\label{fig:icl_control}
\myvspace{-2mm}
\end{figure*}

The auxiliary contexts $c^k$ serve as reliability probes rather than supervision targets (\S\ref{sec:framework}), and the \method student runs zero-shot at inference.
To compare inference-time context with parameter updates, we evaluate the raw model prompted at inference time with the same context-construction strategies used during training (retrieved few-shot, random few-shot, induced instructions; same number of contexts), with no weight updates (Figure~\ref{fig:icl_control}).
Prompting produces task-dependent results: retrieved, topic-matched neighbors yield 86.2 on ScienceQA, compared with 81.7 for random demonstrations; few-shot HumanEval scores are 76.2/73.8 vs.\ 80.5 for Raw.
On ToolAlpaca, the best prompted score is 67.6, compared with 77.9 for the distilled student.
\methodfull (73.3 overall) outperforms the best prompted-teacher configuration (69.0) by 4.3 points and exceeds the per-benchmark best prompted result on five of six benchmarks, despite using no context at inference.
Even where prompting is effective, internalizing the behavior is preferable at deployment: few-shot contexts add roughly 500--1{,}500 prompt tokens per query, whereas the distilled student is evaluated zero-shot on both ID and OOD inputs.

% Data (as a reference; accuracy, not gain)
% Method                      ScienceQA  MBPP  CoS-E  ToolAlpaca  GPQA  HumanEval  Overall
% Raw (zero-shot)             81.5       70.8  81.9   61.8        31.0  80.5       67.9
% Prompted, retrieved FS      86.2       71.6  80.8   66.2        32.8  76.2       69.0
% Prompted, random FS         81.7       72.0  80.9   67.6        31.0  73.8       67.8
% Prompted, induced instr.    79.3       71.6  81.0   67.6        32.8  81.7       69.0
% UniSD* (zero-shot)          85.0       74.7  82.2   77.9        36.4  83.5       73.3

\subsection{Baseline Generalization Across Data Splits}
\label{app:train_split}

\begin{figure*}[t]
\centering
\includegraphics[width=0.95\widefigurewidth]{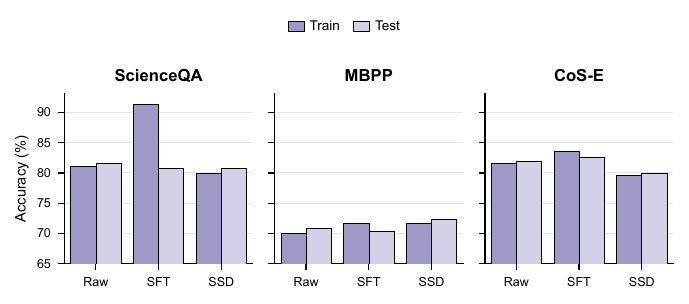}
\caption{Train- and test-split accuracy for SFT, SSD, and Raw (2{,}000 training examples). SFT gains $10.1$ points over Raw on the ScienceQA training split with a test-split score below Raw. SSD is at or slightly below Raw on both splits. These scores distinguish the train--test patterns of the two baselines.}
\label{fig:train_split}
\end{figure*}

We compare SFT, SSD, and Raw on the training split of datasets where the adapted baseline scores are below Raw on the test split (Figure~\ref{fig:train_split}).
SFT scores 91.3 vs.\ 81.2 for Raw on the ScienceQA training split ($+10.1$), with its test-split score below Raw. These results characterize the separation between demonstration fit and test performance.
SSD instead sits at or slightly below raw on both splits, exhibiting a different train--test pattern from SFT.
The comparison uses the shared training configuration in Appendix~\ref{app:exp_setup}, while baselines retain their own objectives and supervision pathways. It characterizes the resulting train--test patterns.
The loss trajectories complement this split-level analysis by showing the observed optimization behavior (Figure~\ref{fig:training_time_component_loss}, right).

% Data (as a reference; train / test)
% Method  ScienceQA     MBPP         CoS-E
% Raw     81.2 / 81.5   70.0 / 70.8  81.7 / 81.9
% SFT     91.3 / 80.8   71.7 / 70.4  83.5 / 82.6
% SSD     80.0 / 80.8   71.7 / 72.4  79.7 / 79.9

\subsection{Robustness to Corrupted Privileged Information}
\label{app:priv_noise}

\begin{figure}[t]
\centering
\includegraphics[width=\singlefigurewidth]{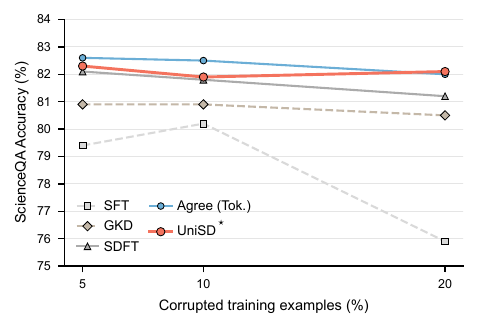}
\caption{Robustness to corrupted privileged information on ScienceQA under 5--20\% corruption. For demonstration-conditioned methods (solid lines) the corruption flips the gold answer shown to the teacher; for SFT and GKD (dashed lines), whose supervision has no privileged channel, it reaches the training targets instead. Relative to clean-data (0\%) accuracy, the reliability-aware variants stay within 0.8 points at every ratio, while SFT drops by 5.2 points.}
\label{fig:priv_noise}
\end{figure}

We test robustness to corrupted privileged information by modifying the demonstrations shown to the self-distillation teacher.
For a fraction of training examples, the gold answer embedded in the teacher's demonstration and retrieved few-shot references is flipped to an incorrect option, so the teacher is misled \emph{before} producing supervision, while the student's prompt and all evaluation data remain clean.
We perturb 5--20\% of the ScienceQA training examples under the same 2{,}000-example budget, one epoch, and canonical hyperparameters, retraining every method from the raw model at each ratio and comparing against each method's clean-data (0\%) accuracy.
SFT and GKD do not receive supervision from a demonstration-conditioned teacher, so for these methods the corruption instead reaches the training targets; we include them as cross-channel references.
Clean-data reference accuracies in this diagnostic are 81.1/81.7/81.6/82.4/82.7 for SFT/GKD/SDFT/Agree (Tok.)/\methodfull, respectively; these differ from the main-table runs. Figure~\ref{fig:priv_noise} shows that the demonstration-conditioned methods remain stable: Agree (Tok.) and \methodfull stay within 0.8 points of their clean-data accuracy at every ratio and remain the top two methods throughout.
The teacher scores the student's on-policy trajectory rather than using the corrupted answer text as a direct training target, and agreement weighting can reduce the influence of context-sensitive supervision.
This control measures the robustness of each complete method under its specified corruption pathway; component-removal experiments follow below.
SFT loses $5.2$ points when corruption instead enters through its training targets, providing a reference under a different supervision pathway.

% Data (as a reference); the 0% (clean-data) column is not plotted.
% Method        Channel                   0%    5%    10%   20%   D(20-0)
% SFT           training target           81.1  79.4  80.2  75.9  -5.2
% GKD           training target (mixed)   81.7  80.9  80.9  80.5  -1.2
% SDFT          teacher's priv. answer    81.6  82.1  81.8  81.2  -0.4
% Agree (Tok.)  teacher's priv. answer    82.4  82.6  82.5  82.0  -0.4
% UniSD*        teacher's priv. answer    82.7  82.3  81.9  82.1  -0.6

\subsection{Leave-One-Out Component Ablation}
\label{app:loo}

\begin{figure}[t]
\centering
\includegraphics[width=\singlefigurewidth]{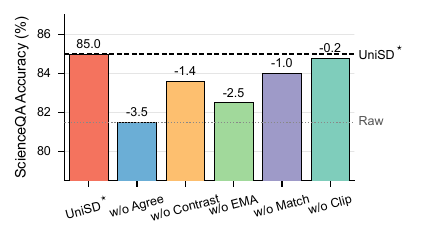}
\caption{Leave-one-out ablation of \methodfull on ScienceQA. Bars show accuracy after removing one component; labels give changes from the full recipe (dashed line). Removing agreement costs $3.5$ points, returning to Raw (dotted line). Removing EMA, contrastive learning, or feature matching costs $1.0$--$2.5$ points; removing clipping costs $0.2$.}
\label{fig:loo_components}
\end{figure}

\begin{figure*}
\centering
\includegraphics[width=0.98\widefigurewidth]{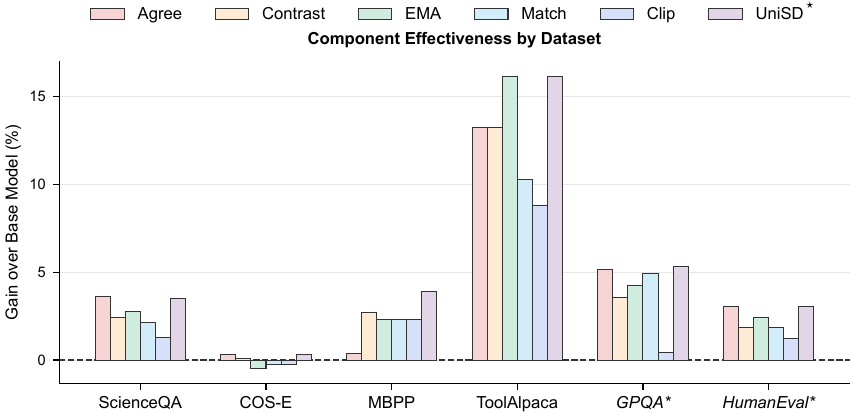}
\caption{
Per-dataset gains of \method variants over the raw Qwen2.5-7B model (disaggregating the middle panel of Figure~\ref{fig:training_time_component_loss}). Asterisks ($\ast$) denote OOD benchmarks. Gains are percentage-point differences and vary by task. CoS-E changes are small: the main table includes a tie for sequence-level agreement and lower scores for EMA, feature matching, and clipping. These observations support task-dependent benefits, while \methodfull gains more strongly on ToolAlpaca ($+16.1$).
}
\label{fig:component_effectiveness}
\end{figure*}

Complementing the standalone results (Table~\ref{tab:overall_qwen25-7b}, Figure~\ref{fig:component_effectiveness}), leave-one-out ablations measure each control's contribution within the full recipe.
We therefore remove exactly one component from \methodfull at a time while keeping all other settings fixed: agreement reweighting ($\gamma{=}0$), token-level contrastive learning (contrastive weight $0$), the EMA teacher (teacher frozen at its initial snapshot), final-layer feature matching (matching weight $0$), and divergence clipping (clip threshold removed).
The leave-one-out experiment uses a 2{,}000-example ScienceQA budget and a full-recipe reference accuracy of 85.0\%.
Figure~\ref{fig:loo_components} shows that the full recipe has the highest accuracy among these configurations.
Agreement reweighting makes the largest conditional contribution: removing it reduces accuracy by $3.5$ points to $81.5$, matching the raw-model reference.
Removing the EMA teacher costs $2.5$ points, and removing contrastive learning or feature matching costs $1.4$ and $1.0$ points, respectively.
Removing divergence clipping changes accuracy by $-0.2$ points, consistent with its role as a numerical safeguard.
Each accuracy change is conditional on the remaining controls being enabled.
Agreement yields the largest leave-one-out effect, followed by EMA, contrastive learning, and feature matching; clipping contributes $0.2$ points under this protocol.

% Data (as a reference; outputs/loo in UniSD_dev, rebuttal E14 rerun)
% Variant        ScienceQA  Delta
% UniSD*         85.0       --      (E13 noise0 reference, checkpoint-250)
% w/o Agree      81.5       -3.5    (gamma_agreement = 0)
% w/o Contrast   83.6       -1.4    (contrastive_weight = 0)
% w/o EMA        82.5       -2.5    (ref_model_mixup_beta = 1, frozen teacher)
% w/o Match      84.0       -1.0    (final_layer_distill_weight = 0)
% w/o Clip       84.8       -0.2    (jsd_token_clip = 0)
% Raw            81.5

\subsection{Additional-Training Control}
\label{app:compute_matched}

\begin{figure*}[t]
\centering
\includegraphics[width=\widefigurewidth]{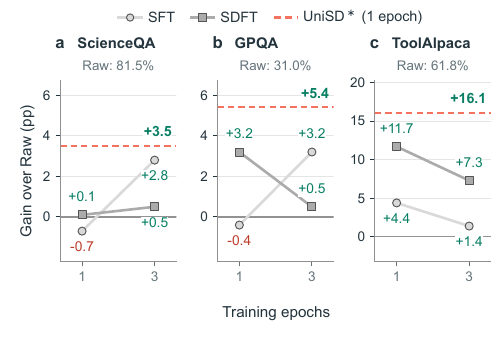}
\caption{Additional-training control on ScienceQA (\textbf{a}), GPQA (\textbf{b}), and ToolAlpaca (\textbf{c}).
Circles/squares denote SFT/SDFT at one and three epochs; dashed lines denote one-epoch \methodfull.
Labels show gains over Raw. ToolAlpaca uses a wider vertical scale.}
\label{fig:compute_matched}
\end{figure*}

% Data (as a reference)
% Method            Budget  ScienceQA  GPQA  ToolAlpaca
% Raw               --      81.5       31.0  61.8
% SFT (1 epoch)     1x      80.8       30.6  66.2
% SFT (3 epochs)    3x      84.3       34.2  63.2
% SDFT (1 epoch)    1x      81.6       34.2  73.5
% SDFT (3 epochs)   3x      82.0       31.5  69.1
% UniSD*            ~2.6x   85.0       36.4  77.9

We test the effect of a longer optimization schedule by increasing the number of baseline training epochs.
We retrain SFT and SDFT for three epochs while keeping their other settings unchanged (Figure~\ref{fig:compute_matched}).
SFT reaches 84.3/34.2/63.2 on ScienceQA/GPQA/ToolAlpaca, and SDFT reaches 82.0/31.5/69.1, versus 85.0/36.4/77.9 for \methodfull.
Under these training schedules, the one-epoch integrated recipe retains an accuracy advantage over both three-epoch baselines.
Training duration is controlled in epochs in this experiment. Separately, Figure~\ref{fig:resource_comparison} reports estimated energy per training token, with a 2.6-times ratio between \methodfull and EMA; that ratio refers to a different pair of methods and cost measure.

\subsection{Sensitivity to the Number of Auxiliary Views}
\label{app:sensitivity}

\begin{figure*}[htbp]
\centering
\includegraphics[width=0.49\widefigurewidth]{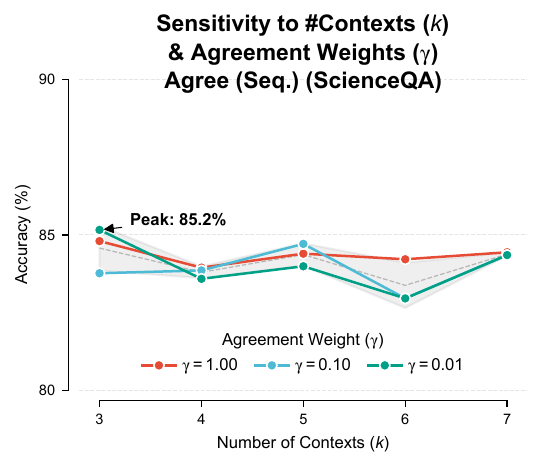}
\includegraphics[width=0.49\widefigurewidth]{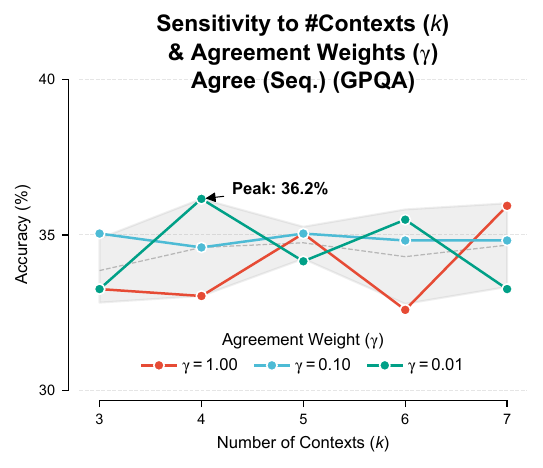}
\myvspace{-3mm}
\caption{Sensitivity to the number of contexts $k$ and the agreement weight 
$\gamma$. The preferred context count depends on the task and agreement strength.}
\label{fig:sensitivity_sequence_qwen25-7b}
\end{figure*}

\begin{figure*}[t]
\centering

\includegraphics[width=0.49\widefigurewidth]{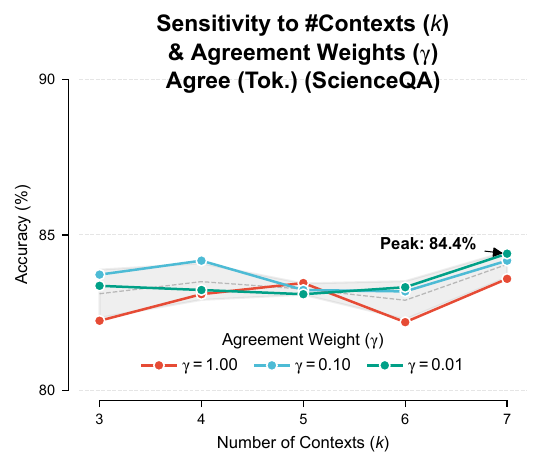}
\includegraphics[width=0.49\widefigurewidth]{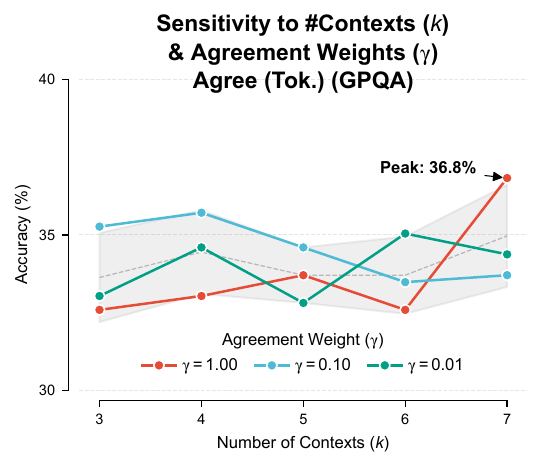}
\caption{Sensitivity to the number of contexts $k$ and the agreement weight $\gamma$. The preferred context count depends on the task and agreement strength.
}
\label{fig:sensitivity_token_qwen25-7B}
\myvspace{-2mm}
\end{figure*}

% Moved from the main-text agreement analysis to fit the ACL page limit.
The sensitivity analyses in Figures~\ref{fig:sensitivity_sequence_qwen25-7b} and \ref{fig:sensitivity_token_qwen25-7B} show that performance changes non-monotonically with $K$.
The best setting depends on both task and granularity: token-level agreement peaks at $K=3$ on ScienceQA with $\gamma=0.01$ (85.2), and at $K=4$ on GPQA with $\gamma=0.01$ (36.2), while sequence-level agreement peaks at $K=7$ on ScienceQA with $\gamma=0.01$ (84.4) and on GPQA with $\gamma=1.0$ (36.8).
The non-monotonic curves motivate selecting complementary, task-relevant auxiliary views. Agreement weights depend on both the context count and the consistency of task-relevant information across views. These findings align with prior studies of context composition and information use~\cite{petronicontext,liu2024lost}.

% ScienceQA
% nctx     3.0    4.0    5.0    6.0    7.0
% gamma                                   
% 1.00   84.80  83.95  84.40  84.22  84.44
% 0.10   83.77  83.86  84.71  82.96  84.35
% 0.01   85.16  83.59  83.99  82.96  84.35

% GPQA
% nctx     3.0    4.0    5.0    6.0    7.0
% gamma                                   
% 1.00   33.26  33.04  35.04  32.59  35.94
% 0.10   35.04  34.60  35.04  34.82  34.82
% 0.01   33.26  36.16  34.15  35.49  33.26

% ScienceQA
% nctx     3.0    4.0    5.0    6.0    7.0
% gamma                                   
% 1.00   82.24  83.09  83.45  82.19  83.59
% 0.10   83.72  84.17  83.23  83.18  84.17
% 0.01   83.36  83.23  83.09  83.32  84.40

% GPQA
% nctx     3.0    4.0    5.0    6.0    7.0
% gamma                                   
% 1.00   32.59  33.04  33.71  32.59  36.83
% 0.10   35.27  35.71  34.60  33.48  33.71
% 0.01   33.04  34.60  32.81  35.04  34.38

\subsection{Auxiliary-Context Construction}
\label{app:context_construction}

\begin{figure*}[ht]
\centering
\includegraphics[width=0.98\widefigurewidth]{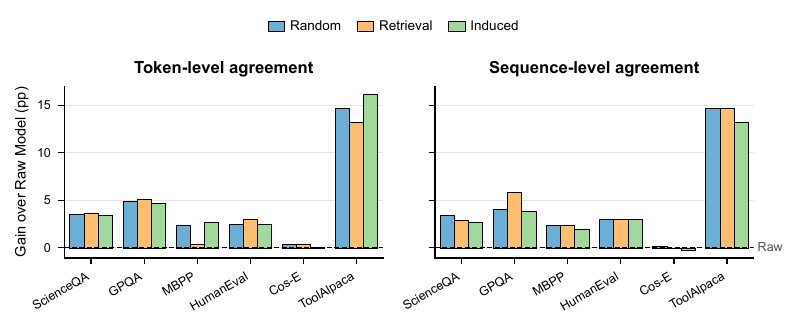}
\caption{Gains over the raw Qwen2.5-7B model (dashed line) across six benchmarks for the three auxiliary-context construction strategies --- random, retrieval, and induced --- under token-level (left) and sequence-level (right) agreement.}
\label{fig:radar_token_qwen25-7b}
\end{figure*}

Figure~\ref{fig:radar_token_qwen25-7b} compares the auxiliary-context strategies analyzed in Section~\ref{sec:agree}.

% Data (as a reference)

% Method	ScienceQA	GPQA	MBPP	HumanEval	Cos-E	ToolAlpaca
% Raw	81.52	31.03	72.37	80.49	81.9	61.76
% Random	85.07	35.94	73.15	82.93	82.31	76.47
% Retrieval	85.16	36.16	71.21	83.54	82.23	75.0
% Induced	84.94	35.71	73.54	82.93	81.98	77.94

% Queue the two-column figure before the next page is assembled.
\begin{figure*}[t]
\centering
\includegraphics[width=\widefigurewidth]{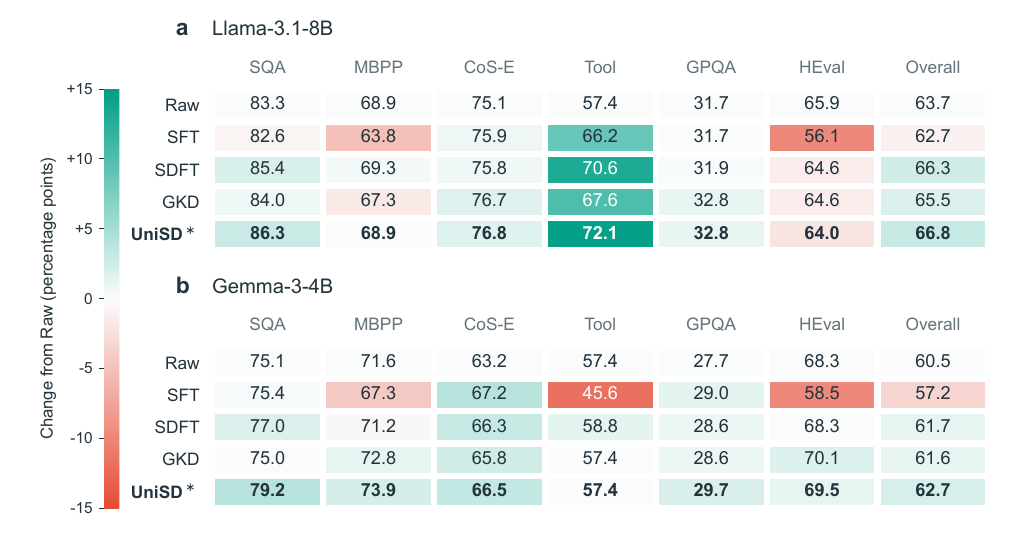}
\caption{Absolute scores and changes relative to Raw for \textbf{a,} Llama-3.1-8B and \textbf{b,} Gemma-3-4B.
Cell labels give absolute percentages ($\uparrow$); color encodes the change from Raw on a shared scale, with green/red denoting higher/lower accuracy.
Abbreviations follow Table~\ref{tab:overall_qwen25-7b}. Overall is the six-benchmark mean.}
\label{fig:cross_family}
\end{figure*}

\subsection{Effect of Model Scale}
\label{app:model_scale}

% figure* == figure in one-column layouts; spans both columns in the ACL build.
\begin{figure*}[t]
\centering
\includegraphics[width=0.62\widefigurewidth]{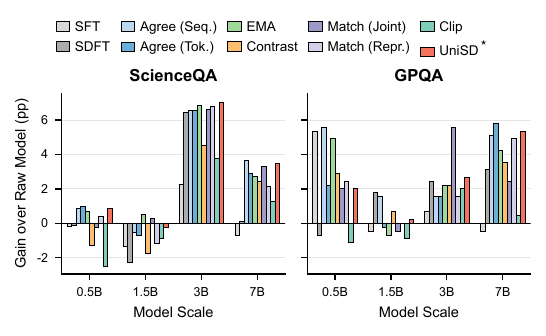}
\includegraphics[width=0.31\widefigurewidth]{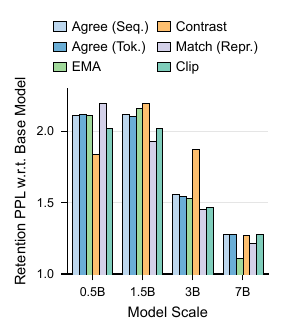}
\caption{\textbf{Left.} Gains over the raw Qwen2.5 model~\citep{qwen25} across four size variants on ScienceQA (in-domain) and GPQA (OOD). \methodfull reaches its largest ScienceQA gain (+7.06 percentage points) at 3B. \textbf{Right.} Base-distribution retention perplexity across the same Qwen2.5 size variants.
}
\label{fig:gain_and_retention_ppl_qwen25}
\myvspace{-4mm}
\end{figure*}

Figure~\ref{fig:gain_and_retention_ppl_qwen25} shows task gains and base-distribution retention perplexity across the four Qwen2.5 sizes, complementing the analysis in \S\ref{sec:generalization_models}.

\subsection{Absolute Cross-Family Results}
\label{app:cross_family_absolute}

Figure~\ref{fig:cross_family} supplies the absolute Llama-3.1-8B and Gemma-3-4B scores underlying the cross-family analysis.
The Qwen2.5-7B results appear in Table~\ref{tab:overall_qwen25-7b}.
All six-benchmark means use the same task set.
Llama GKD achieves 67.6\% on ToolAlpaca and a six-benchmark mean of 65.5\%.

\subsection{Far-OOD Capability Preservation}
\label{app:far_ood}

\begin{figure*}[t]
\centering
\includegraphics[width=0.9\widefigurewidth]{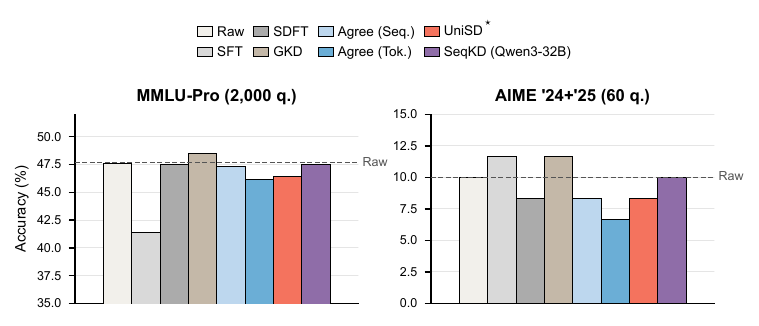}
\caption{Far-OOD evaluation of ScienceQA-trained checkpoints on MMLU-Pro~\cite{wang2024mmlu} and AIME 2024--2025~\cite{aime25}. On MMLU-Pro, SFT loses 6.2 points while the self-distillation variants stay within 1.4 points of Raw. On AIME, \methodfull solves 5/60 problems versus Raw's 6/60.}
\label{fig:far_ood}
\end{figure*}

% Data (as a reference)
% Method             MMLU-Pro (2000q)  AIME'24+'25 (60q)
% Raw                47.6              10.0
% SFT                41.4              11.7
% SDFT               47.5               8.3
% GKD                48.5              11.7
% Agree (Seq.)       47.3               8.3
% Agree (Tok.)       46.2               6.7
% UniSD*             46.4               8.3
% SeqKD Qwen3-32B    47.5              10.0

This probe evaluates ScienceQA-trained checkpoints on 2{,}000 MMLU-Pro questions and 60 AIME 2024--2025 problems.
\methodfull scores 46.4\% on MMLU-Pro versus 47.6\% for Raw and 41.4\% for SFT.
On AIME, its 8.3\% corresponds to 5/60 correct answers, versus 6/60 (10.0\%) for Raw.
These scores quantify distant capability retention after ScienceQA adaptation, with AIME reported as both accuracy and correct-answer counts.

% Queue the wide dataset table before its section starts on the next page.
% table* == table in one-column layouts; spans both columns in the ACL build.
\begin{table*}[t]
\centering
\small
\setlength{\tabcolsep}{2.5pt}
\renewcommand{\arraystretch}{1.08}
\begin{tabular*}{\textwidth}{@{\extracolsep{\fill}}llrrl}
\toprule
Category & Dataset & Train & Eval. & Declared license \\
\midrule
Scientific Reasoning
  & \href{https://github.com/lupantech/ScienceQA}{ScienceQA} & 12,726 & 4,241 & CC BY-NC-SA 4.0 \\
  & \href{https://huggingface.co/datasets/Idavidrein/gpqa}{GPQA} & & 448 & CC BY 4.0 \\
\addlinespace[2pt]
Coding
  & \href{https://huggingface.co/datasets/google-research-datasets/mbpp}{MBPP} & 120 & 257 & CC BY 4.0 \\
  & \href{https://huggingface.co/datasets/openai/openai_humaneval}{HumanEval} & & 164 & MIT \\
\addlinespace[2pt]
Commonsense QA
  & \href{https://github.com/salesforce/cos-e}{CoS-E} & 9,741 & 1,221 & BSD-3-Clause \\
\addlinespace[2pt]
Tool Usage
  & \href{https://github.com/tangqiaoyu/ToolAlpaca}{ToolAlpaca} & 4,046 & 68 & Apache-2.0 \\
\addlinespace[2pt]
General Knowledge
  & \href{https://huggingface.co/datasets/TIGER-Lab/MMLU-Pro}{MMLU-Pro} & & 2,000 & MIT \\
\addlinespace[2pt]
Competition Mathematics
  & \href{https://huggingface.co/datasets/HuggingFaceH4/aime_2024}{AIME 2024} & & 30 & Apache-2.0 (upstream) \\
  & \href{https://huggingface.co/datasets/math-ai/aime25}{AIME 2025} & & 30 & Apache-2.0 \\
\bottomrule
\end{tabular*}
\caption{Training and evaluation sample counts used in this paper and provider-declared licenses. Blank training cells indicate evaluation-only benchmarks. Dataset names link to source repositories or cards. For AIME 2024, the license is declared by its cited upstream source; see Appendix~\ref{app:dataset_licenses} for details.}
\label{tab:dataset_stats}
\end{table*}

\section{Additional Experimental Details}
\label{app:exp_setup}

\paragraph{Training Configuration.}
We use the LoRA and AdamW settings in \S\ref{sec:exp_setup}, with optimizer coefficients $\beta_1=0.9$ and $\beta_2=0.999$.
Unless otherwise noted, training runs for one epoch with gradient accumulation of four steps and bf16 mixed precision.
On-policy completions are generated with vLLM~\cite{kwon2023efficient} in colocate mode.
The maximum prompt and completion lengths are $3072$ and $1024$ tokens, respectively.
The standard \method training entry point disables intermediate evaluation and saves the checkpoint at epoch end.
Experiments run on six NVIDIA A100 80GB GPUs.
The EMA-retention sweep and disagreement diagnostic specify their run settings, sample counts, and protocols in Appendices~\ref{app:ema_retention} and \ref{app:disagreement_correctness}.

\subsection{Supervision and Context Construction}
\label{app:supervision_construction}

\paragraph{Primary teacher.}
The task-specific demonstrations supplied to the primary teacher (\S\ref{sec:preliminary}) contain gold answers and rationales for ScienceQA and CoS-E, reference code for MBPP, and reference tool calls with associated demonstration text for ToolAlpaca.

\paragraph{Auxiliary reliability views.}
We retrieve the auxiliary examples defined in \S\ref{sec:framework} using cosine similarity between \texttt{all-MiniLM-L6-v2} embeddings of questions or task prompts, excluding the current example.
The context-construction ablation substitutes random training examples, again excluding the current example, or cached instructions induced from training demonstrations.

\paragraph{Contrastive negatives.}
The same backbone is prompted for four candidate negatives per example at temperature 1.0, with structured reasoning and answer fields.
Multiple-choice prompts specify a non-gold option; coding prompts request a bug; tool-use prompts request an incorrect action, arguments, interpretation, or extra call.
One cached candidate is selected using the run seed when constructing the training dataset.
Candidate labels and inclusion are determined by the generation prompts, and candidates are cached without execution-based filtering.
Cached answer labels must remain consistent with each example's option ordering.

\paragraph{Objective implementation.}
The fixed objective configuration is specified in Appendix~\ref{app:objective_spec}.
The implementation selects forward KL, $D_{\mathrm{KL}}(\pi^{\mathrm T}\|\pi_\theta)$, with its $\alpha=0$ option; baselines retain their own objectives.
Feature matching uses completion-token final-layer hidden states, while the primary KL term supplies logit matching.
The context-count, agreement-strength, reliability-constant, and EMA-retention sweeps use separately identified configurations.

\paragraph{Evaluation.}
We use vLLM with the decoding settings and metrics in \S\ref{sec:exp_setup}.
Multiple-choice answers are extracted automatically; MBPP and HumanEval programs are evaluated with sandboxed tests and a 10-second timeout.

\subsection{Dataset Licenses and Release Scope}
\label{app:dataset_licenses}

Table~\ref{tab:dataset_stats} records the licenses declared by the linked source repositories and dataset cards.
We use English-language benchmark text; MBPP and HumanEval additionally contain Python programs.
For GPQA, CC BY 4.0 covers the dataset, while MIT covers the accompanying code; its access terms ask users not to publish question text or images online.

The AIME probe combines 30 problems from \texttt{HuggingFaceH4/aime\_2024} and 30 from \texttt{math-ai/aime25}.
The AIME 2024 card points to \href{https://huggingface.co/datasets/AI-MO/aimo-validation-aime}{\texttt{AI-MO/aimo-validation-aime}}, whose distribution declares Apache-2.0; the H4 subset does not state a separate license.
The AIME 2025 distribution also declares Apache-2.0.
These distribution labels do not transfer copyright in the original examination problems.

Our planned release contains implementation code only, without datasets, cached examples, generated samples, or model weights.
Users obtain data and base models from their providers under the applicable licenses and access terms.

\paragraph{Model and software licenses.}
Qwen2.5-0.5B/1.5B/7B-Instruct use Apache-2.0; Qwen2.5-3B-Instruct uses the \href{https://huggingface.co/Qwen/Qwen2.5-3B-Instruct/blob/main/LICENSE}{Qwen Research License}.
Llama-3.1-8B-Instruct uses the \href{https://huggingface.co/meta-llama/Llama-3.1-8B-Instruct/blob/main/LICENSE}{Llama 3.1 Community License}, and Gemma-3-4B-IT uses the \href{https://ai.google.dev/gemma/terms}{Gemma Terms of Use}.
The \texttt{all-MiniLM-L6-v2} embedding model, Transformers, TRL, and vLLM use Apache-2.0; PyTorch uses the BSD-3-Clause license.
Our implementation uses Apache-2.0 and documents environment setup, training modes, and preparation of auxiliary caches.

\section{Broader Impact}
\label{app:impact}

\method explores self-distillation as a way for LLMs to improve using supervision derived from their own behavior, rather than relying on stronger external teachers. 
This may lower the cost and access barriers to post-training, especially for academic groups, smaller organizations, and resource-constrained settings.
It can also reduce the need to transmit in-domain data to external models, which makes the approach appealing for privacy-sensitive or local adaptation. 
Finally, \method provides a unified, extensible, reproducible and controllable framework for studying self-distillation.

\ifdefined\aidisclosureinmain
\else
\section{Use of AI Assistants}
\label{app:gpt}
AI assistants were used as auxiliary tools in preparing this manuscript, primarily for language refinement, clarity, organization, and limited experimental workflows.
The experimental design and methodological choices were made by the authors. All results, analyses, and final content were manually checked and verified by the authors.

\fi

% For our \textbf{agreement-based variants}, we use agreement strength $\gamma \in \{0.01, 0.1, 1.0\}$ and number of auxiliary contexts $n_{\text{ctx}} \in \{3,5,7\}$.
% For \textbf{EMA}, we search sync steps in $\{1,10,100\}$ and mixing coefficient $\alpha \in \{0.01,0.1,0.5\}$.
% For \textbf{token-level contrastive learning}, we search over contrastive loss weight $\in \{0.01,0.1,1.0\}$ and margin $\in \{0.01,0.1,1.0\}$.
% For \textbf{final-layer distillation}, we search over distillation weight $\in \{0.01,0.1,1.0\}$.

% For Qwen3-8B~\cite{qiu2025gated}, we disabled thinking mode during training and evaluation. During training, thinking mode introduces substantial \texttt{<think>} tokens into completions, which are not part of the target answer but are still included in the KL objective, thereby weakening the effective distillation signal. During evaluation, thinking mode also interferes with output parsing, since our harnesses for multiple-choice QA, code generation, and tool-use tasks expect answers in a strict format such as an option letter, a code block, or a JSON tool call.

% 1. 污染 completion tokens — 训练时模型生成的 completion 里会夹带大量
%   <think> 内容，而这些 token 不是我们要蒸馏的目标答案，却会被算进 KL
%   loss，浪费训练信号。
% 2. 破坏 eval 解析 — 评估时（MCQA、code、tooluse），harness
%   期望直接输出答案（选项字母、代码块、JSON tool call）。thinking
%   输出会导致提取器匹配不到正确格式，结果全部报错。

% \input{tables/tab-teacher_prompt}

% The conference entry point controls placement of the scope and ethics sections.
\section{Future Work}
\label{app:future}
% Extension to long-horizon agentic tasks where the model needs to complete multiple steps to achieve a goal.
% Improvement in evaluation. For example, models should still learn from partially correct answers in challenging tasks instead of completely discarding them.
% Future work can explore diverse contrastive learning strategies.
% First, extending \method to long-horizon agentic tasks would test whether reliability-aware on-policy distillation remains effective under sparse and delayed feedback.
% Second, evaluation protocols should better account for partially correct answers, rather than treating them as entirely incorrect.
% Finally, richer contrastive objectives and disagreement measures may further improve the quality of self-derived supervision, including connections to recent work on prompt optimization~~\cite{wan2025beyond}

\paragraph{Long-Horizon Agentic Settings.}
Extending \method to long-horizon agentic tasks will examine adaptation across interdependent decisions with sparse, delayed feedback.

\paragraph{Finer-Grained Trajectory Evaluation.}
Finer-grained evaluation could credit partially correct reasoning and useful intermediate steps, complementing our outcome-based protocols and measuring benefits beyond final-answer accuracy.

\paragraph{Broader Self-Supervision Objectives.}
We can combine contrastive learning, cross-view disagreement, and prompt optimization~\cite{wan2025beyond}.

\end{document}